\documentclass{article}

\usepackage{times}
\usepackage{graphicx} % more modern
\usepackage{subcaption} 

\usepackage{amsmath,amssymb}

\usepackage{pgfplots}

\usepackage{natbib}

\usepackage{algorithm}
\usepackage{algorithmic}

\usepackage{hyperref}

\usepackage{dblfloatfix}
\usepackage[accepted]{arxiv}

\icmltitlerunning{Learning Localized Spatio-Temporal Models From Streaming Data}

\DeclareMathOperator{\col}{col}
\DeclareMathOperator{\tr}{tr}
\DeclareMathOperator{\E}{E}
\DeclareMathOperator{\Cov}{Cov}

\newcommand{\mbf}[1]{\mathbf{#1}}
\newcommand{\mbs}[1]{\boldsymbol{#1}}
\newcommand{\what}[1]{\widehat{#1}}
\newcommand{\wtilde}[1]{\widetilde{#1}}

\newcommand{\giv}[1]{\underline{#1}}
\newcommand{\0}{\mbf{0}}
\newcommand{\I}{\mbf{I}}
\newcommand{\1}{\mbf{1}}
\newcommand{\T}{\top}

\newcommand{\GPR}{\textsc{Gpr}}
\newcommand{\SPICE}{\textsc{Spice}}
\newcommand{\mapN}{\texttt{N}}
\newcommand{\mapS}{\texttt{S}}
\newcommand{\mapE}{\texttt{E}}
\newcommand{\mapW}{\texttt{W}}

\newcommand{\dataset}{\mathcal{D}}
\newcommand{\basis}{\mbs{\phi}}
\newcommand{\s}{\mbf{s}}
\newcommand{\y}{\mbf{y}}
\newcommand{\parameter}{\mbs{\theta}}
\newcommand{\pvec}{\parameter}
\newcommand{\covmat}{\mbf{K}}
\newcommand{\samplecovmat}{\wtilde{\covmat}}
\newcommand{\psub}{\theta}
\newcommand{\pmat}{\mbs{\Theta}}
\newcommand{\basismat}{\mbf{\Phi}}

\newcommand{\sbasis}{\mbs{\varphi}}
\newcommand{\tbasis}{\mbs{\psi}}

\newcommand{\tbasissc}{\psi}

\newcommand{\bfunc}{f}
\newcommand{\lin}{\mbs{\lambda}}

\DeclareMathOperator*{\argminA}{\arg\min}

\begin{document} 

\twocolumn[
%\icmltitle{Online Learning of Localized Spatio-Temporal Models}
\icmltitle{Learning Localized Spatio-Temporal Models From Streaming Data}



% you can specify symbols, otherwise they are numbered in order
\icmlsetsymbol{equal}{*}

\begin{icmlauthorlist}
\icmlauthor{Muhammad Osama}{uu}
\icmlauthor{Dave Zachariah}{uu}
\icmlauthor{Thomas B. Sch\"{o}n}{uu}
\end{icmlauthorlist}

\icmlaffiliation{uu}{Uppsala University, Sweden}

\icmlcorrespondingauthor{Muhammad Osama}{muhammad.osama@it.uu.se}
\icmlcorrespondingauthor{Dave Zachariah}{dave.zachariah@it.uu.se}

\icmlkeywords{TODO}

\vskip 0.3in
]



\printAffiliationsAndNotice{}  % leave blank if no need to mention equal contribution

\begin{abstract} 
We address the problem of predicting spatio-temporal processes with temporal patterns that vary across spatial regions, when data is obtained as a stream. That is, when the training dataset is augmented sequentially. Specifically, we develop a localized spatio-temporal covariance model of the process that can capture spatially varying temporal periodicities in the data. We then apply a covariance-fitting methodology to learn the model parameters which yields a predictor that can be updated sequentially with each new data point. The proposed method is evaluated using both synthetic and real climate data which demonstrate its ability to accurately predict data missing in spatial regions over time.
\end{abstract} 

\section{Introduction}

\begin{figure}[!h]
  \begin{center}
    \includegraphics[width=0.450\columnwidth]{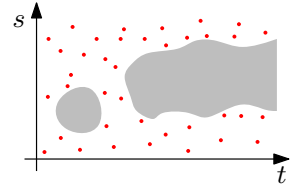}
  \end{center}
  %\vspace{0cm}
  \caption{Example of training points (dots) in $\dataset_n$ over a bounded space-time domain $\mathcal{S} \times \mathcal{T}$ with one spatial dimension. Note that the sampling pattern may be irregular and that it is not possible to provide complete spatial coverage at all times. A typical problem is to predict the process in unobserved regions (shaded).}
  \label{fig:problemsetup}
\end{figure}

Many real-world processes of interest, ranging from climate variables to brain signals, are spatio-temporal in nature, cf. \citet{cressie2011statistics}. That is, they can be described as a random quantity that varies over some fixed spatial and temporal domain. Suppose we obtain $n$ training points from a real-valued spatio-temporal process,
\begin{equation*}
\dataset_n = \big\{ \: (\s_1, t_1, y_1), \: \dots, \: (\s_n, t_n, y_n) \: \big\},
\end{equation*}
where $y_i$ denotes the quantity of interest observed at the $i^\text{th}$ training point, with spatial coordinate $\s_i$ and time $t_i$. For notational convenience, let $(\s,t,y)$ denote an unobserved test point in space-time where $y$ is unknown. Then a common goal is to predict $y$ in unobserved space-time regions $(\s,t)$ using $\dataset_n$. Specifically, certain spatial regions may have limited data coverage over extended periods of time, as illustrated in Figure~\ref{fig:problemsetup}.
In real-world applications, $\dataset_n$ need not be gathered in a single batch but obtained in parts over time from various sensors, stations, satellites, etc. That is, the dataset  is augmented sequentially, i.e., $n=1, 2, \dots, N$. In these streaming data scenarios, we are interested in continuous refinement of the prediction of $y$ at $(\s,t)$ as new data is augmented into $\dataset_{n+1}$.

The unknown data-generating process is often assumed to belong to a class of data models indexed by a parameter $\pvec$. Each model $\pvec$ in the class yields a predictor $\what{y}_{\psub}(\s, t)$ of $y$ at test point $(\s,t)$. A specific set of model parameters $\what{\pvec}$ is learned using $\dataset_n$. Examples of commonly used model classes include Gaussian Processes (GP) \cite{rasmussen2006gaussian}, spatio-temporal random effects models \cite{cressie2010fixed}, dynamic factor analysis models \cite{lopes2008spatial,fox2015bayesian}, spatial random effect models extended to incorporate time as an additional dimension \cite{zammit2017frk} (cf. related work section below). For many spatio-temporal applications, the model class should be capable of expressing temporal patterns that change across different spatial regions. Moreover, for streaming data scenarios, the learned parameter $\what{\pvec}$ and the resulting predictor $\what{y}_{\what{\psub}}(\s, t)$ should be updated in a sequential manner.

Our contribution in this paper is two-fold:
\begin{itemize}
    \item we develop a non-stationary, localized covariance model capable of capturing temporal patterns that change across space, as illustrated in Figure~\ref{covstr} below.
    
    \item we show how to sequentially learn the covariance model parameters and update the predictor from streaming spatio-temporal data, with a runtime that is linear in $n$.
\end{itemize}
In Section~\ref{sec:review}, we relate our work to already existing approaches and introduce a commonly used model class in Section~\ref{sec:modelclass}. In Section~\ref{sec:proposed} we develop a localized spatio-temporal covariance model to be used in conjunction with a covariance-fitting learning approach. Finally, the proposed method is evaluated using synthetic and real climate data in Sections~\ref{sec:synthetic} and \ref{sec:realdata}, respectively.

\textbf{Notation:} $\col\{ \s_1, \s_2 \}$ stacks both elements into a single column vector. $\otimes$, $\delta(\cdot)$, $\| \cdot \|_{\mbf{W}}$ and $\dagger$ denote the Kronecker product, Kronecker delta function, weighted $\ell_2$-norm and Moore-Penrose inverse, respectively. Finally, the sample mean is denoted by $\what{\E}[\s_i] = \frac{1}{n} \sum^n_{i=1} \s_i$.

\section{Related work}
\label{sec:review}

A popular model class is the family of GPs, specified by a mean and covariance function \cite{rasmussen2006gaussian}. This approach is computationally prohibitive in its basic form since both learning the model parameters $\pvec$ and implementing the predictor $\what{y}_{\psub}(\s,t)$ requires a runtime on the order of $\mathcal{O}(N^3)$, where $N$ is typically large in spatio-temporal applications. The predictor implementation can be approximated using various techniques. One popular approach is to approximate the training data using $m\ll N$ inducing points which reduces the runtime to $\mathcal{O}(m^2N)$ \cite{quinonero2005unifying, bijl2015online}. Moreover, by assuming Kronecker covariance functions it is possible to obtain even shorter runtimes by utilizing the Kronecker structure of the GP covariance matrix \cite{saatcci2012scalable}. If the model class is restricted to stationary covariance functions, the runtimes can be reduced further, cf. \cite{saatcci2012scalable,wilson2014fast}. In the space-time domain, such models are also equivalent to dynamical system models so that $\what{y}_{\psub}(\s, t)$ can be approximated using a basis expansion and implemented by a Kalman smoother \cite{sarkka2013spatiotemporal}. In the above cases, however, $\pvec$ and $\what{y}_{\psub}(\s, t)$ are not updated jointly when obtaining streaming data. 

The restriction to stationary covariance models is, moreover, not always adequate to capture temporal patterns that differ across spatial regions. This modeling limitation is addressed by \citet{cressie2010fixed}, where a discrete-time model class is partially specified using a spatial basis function expansion with time-varying  expansion coefficients. These are modeled as a first-order vector auto-regressive process. The coefficients thus determine a spatial pattern of the process that evolves at each discrete time-instant. This model class can capture patterns localized to specific regions in space, unlike stationary covariance models. The predictor $\what{y}_{\psub}(\s, t)$ can be viewed as a spatial fixed-rank kriging method that is updated via a Kalman filter and thus applicable to streaming data (cf. \citet{cressie2008fixed}). The model parameter $\pvec$, however, is learned using a moment-fitting approach and operates on batch rather than streaming datasets. Other work using dynamic factor analysis models \cite{lopes2008spatial,fox2015bayesian} similarly allow for time-varying coefficients but with more flexible data-adaptive basis. However, they are implemented using Markov Chain Monte Carlo methods which are computationally prohibitive for the scenarios considered herein.

Moreover, a first-order auto-regressive structure may not accurately capture more complex temporal patterns observed in real spatio-temporal processes. The approach taken by \cite{zammit2017frk} circumvents this limitation by basis functions that are localized in both space and time. Time locality cannot, however, capture periodic patterns or trends necessary for interpolation over longer periods. The model parameters are learned using an expectation-maximization method which is not readily applicable to streaming data scenarios.

\section{Spatio-temporal model class}
\label{sec:modelclass}

We begin by defining the data vector $\y = \col\{y_1, y_2, \ldots, y_n\}$ obtained from $\dataset_n$. For the test point $(\s, t)$, we consider the unbiased predictor of $y$ as a linear combination of the data \cite{stein2012interpolation}:
\begin{equation} \label{eq:predictor}
    \what{y}(\s,t) = \lin^\T(\s,t)\y,
\end{equation}
where $\lin^\T(\s,t)$ is a vector of $n$ weights which naturally depend on the test point $(\s,t)$. The weight vector is defined as the minimizer of the conditional mean square prediction error. That is,
\begin{equation}
  \lin(\s,t) \triangleq \argminA_{\giv{\lin}} \: \E\Big[ \; (y-\giv{\lin}^\T\y)^2 \; \big| \; \s,t \; \Big].
\label{eq:optimalweights}
\end{equation}
Since the conditional error is determined by the unknown distribution $p(y,\y|\s,t, \s_1,t_1, \dots, \s_n,t_n )$, we specify a class of data-generating models, using only the mean and covariance \cite{cressie2011statistics}:
\begin{equation} \label{eq:modelclass}
\begin{cases}
\E[y] &= \mbf{u}^\T(\s,t)\mbs{\eta} \\
\Cov[y,y'] &= \basis^\T(\s,t)\pmat\basis(\s',t') + \theta_0\delta(\s,\s')\delta(t,t') .
\end{cases}
\end{equation}
The function $\mbf{u}(\s,t)$ captures the expected trend of the entire spatio-temporal process $y$, and when there is no such general trend we set $\mbf{u}(\s,t) \equiv 1$. The function $\basis(\s,t)$ captures the smoothness of the process in space-time and is of dimension $p\times 1$. The parameter matrix $\pmat$ is diagonal and specifies the relevance of each dimension of $\basis(\s,t)$ similar to the way in which automatic relevance determination is sometimes used within the GP \cite{tipping2001sparse, faul2002analysis}. Taken together, \eqref{eq:modelclass} specifies a class of models, each of which is indexed by the parameters $(\mbs{\eta}, \pmat, \theta_0)$. The $p+1$ covariance parameters $(\pmat, \psub_0)$, which we collectively 

\begin{table*}[!b]
%\caption{Derivation of Result XYZ}
\centering
\begin{minipage}{0.7\textwidth}
\begin{equation}
  \varphi(s) =
  \begin{cases}
   \frac{1}{6}\bfunc(s)^3 & \text{$\frac{(c-2)L}{4}\leq s <\frac{(c-1)L}{4}$}\\
   \frac{-1}{2}\bfunc(s)^3+2\bfunc(s)^2-2\bfunc(s)+\frac{2}{3} & \text{$\frac{(c-1)L}{4}\leq s <\frac{Lc}{4}$}\\
   \frac{1}{2}\bfunc(s)^3-4\bfunc(s)^2+10\bfunc(s)-\frac{22}{3} & \text{$~~~~~~~\frac{Lc}{4}\leq s <\frac{(c+1)L}{4}$}\\
   \frac{-1}{6}\bfunc(s)^3+2\bfunc(s)^2-8\bfunc(s)+\frac{32}{3} & \text{$\frac{(c+1)L}{4}\leq s \leq\frac{(c+2)L}{4}$}\\
   0 & \text{otherwise}
  \end{cases}
\quad \text{where} \quad \bfunc(s) = \frac{4s}{L} - c+2 
\label{eq:bspline}
\end{equation}
\medskip
\hrule
\end{minipage}
\end{table*}

denote by $\pvec = \col\{ \psub_0, \psub_1, \dots, \psub_p \}$ for notational convenience, determine the spatio-temporal covariance structure $\Cov_{\psub}[y,y']$ which depends on the function $\basis(\s,t)$. In the next section, we will specify $\basis(\s,t)$ to develop a suitable covariance model to capture local spatial and periodic temporal patterns.

For a given model in the class, the optimal weights \eqref{eq:optimalweights} are  given in closed form \cite{stein2012interpolation} as 
\begin{equation*}
\begin{split}
\mbs{\lambda}_{\psub}(\s,t)&=\covmat^{-1}\1 (\1^\T\covmat^{-1}\1)^\dagger +\covmat^{-1}\mbf{\Pi}^\perp \basismat \pmat \basis(\s,t),
\end{split}
\end{equation*}
where the subindex highlights the model parameter dependence. The quantities in $\mbs{\lambda}_{\psub}(\s,t)$ are determined by the regressor matrix
\begin{equation*}
    \basismat = \begin{bmatrix} \basis(\s_1,t_1) & \ldots & \basis(\s_n,t_n) \end{bmatrix}^\T
\end{equation*}
and the following covariance matrix
\begin{equation}
\begin{split}
\covmat_\psub &=\Cov[\y,\y]=\basismat \pmat \basismat^\T + \theta_0\mbf{I} \; \succ \; \0 \\
\end{split}
\label{eq:covmat}
\end{equation} 
with $\mbf{\Pi}^\perp = \I - \1 (\1^\T \covmat^{-1} \1)^\dagger \1 \covmat^{-1}$ being an oblique projector onto $\text{span}(\1)^\perp$.

The optimal weights are invariant to the mean parameters  $\mbs{\eta}$ and to uniform scaling of the $p+1$ covariance parameters $\pvec$. By learning $\pvec$ up to an arbitrary scale factor, the predictor \eqref{eq:predictor} is given by the linear combiner weights $\mbs{\lambda}_\psub(\s,t)$. If we assume that the process is Gaussian, the model can be learned using the maximum likelihood framework. However, this yields neither a convex problem nor one that is readily solved in a sequential manner as $\dataset_n$ is augmented sequentially. In the next section, we apply a convex covariance-fitting framework to learn the spatio-temporal model using streaming data.  

\section{Proposed method}
\label{sec:proposed}

\begin{figure*}[ht!]
%\vskip 0.1in
\begin{center}
\begin{subfigure}{0.33\textwidth}
  \centering
  \includegraphics[width=0.9\linewidth]{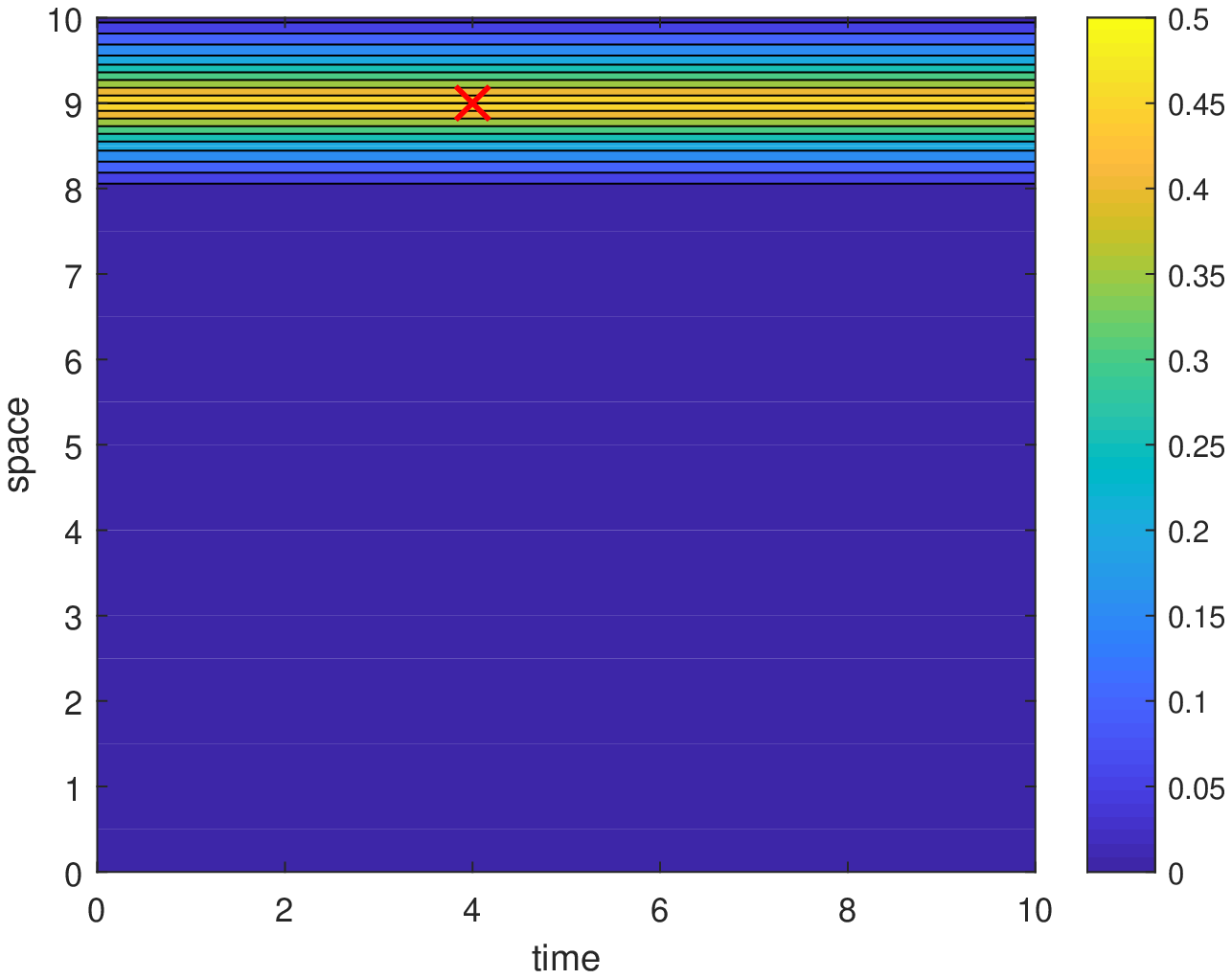}
  \caption{}
  \label{fig:covstr_a}
\end{subfigure}
\hfill
\begin{subfigure}{0.33\textwidth}
  \centering
  \includegraphics[width=0.9\linewidth]{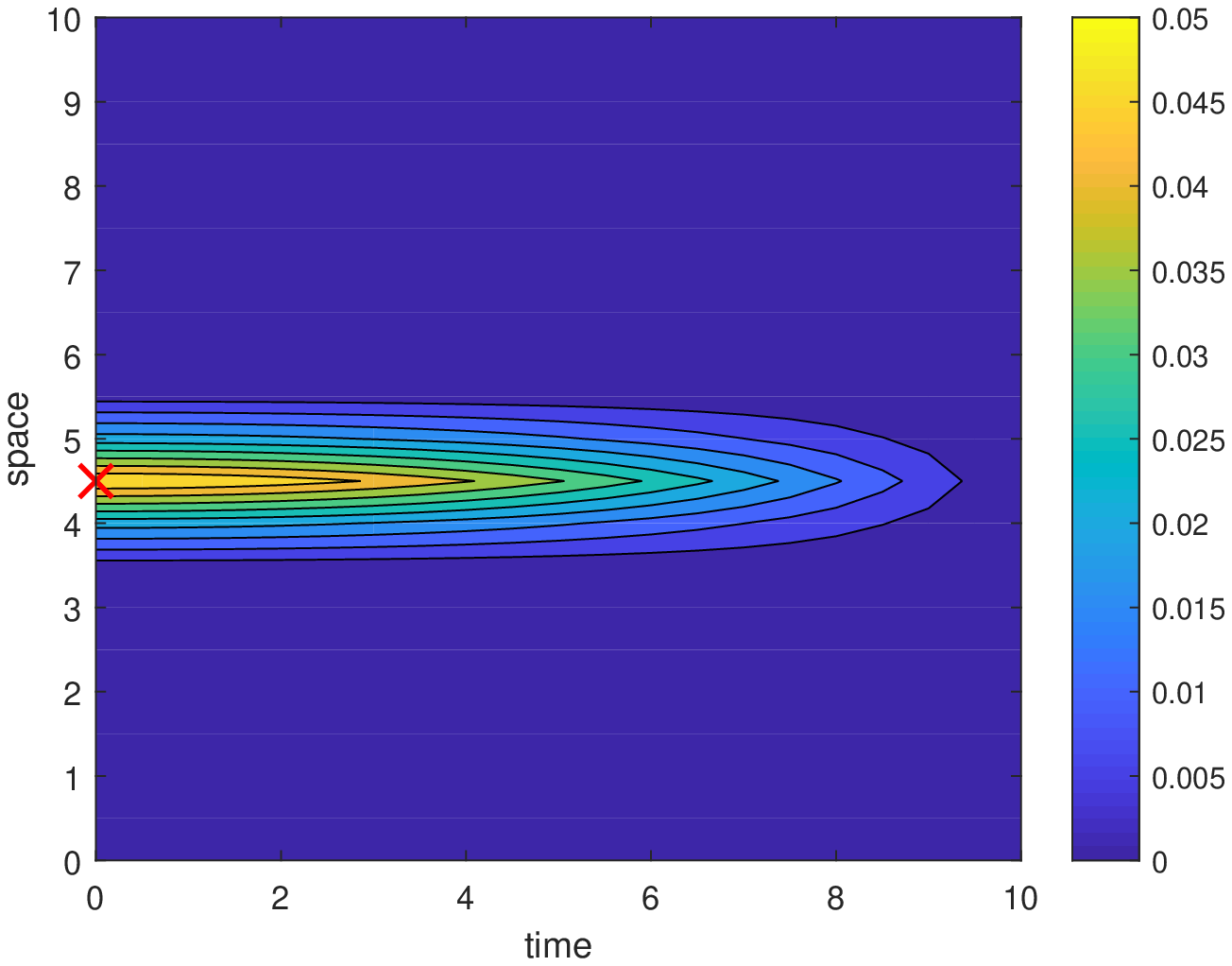}
  \caption{}
  \label{fig:covstr_b}
\end{subfigure}
\begin{subfigure}{0.33\textwidth}
  \centering
  \includegraphics[width=0.9\linewidth]{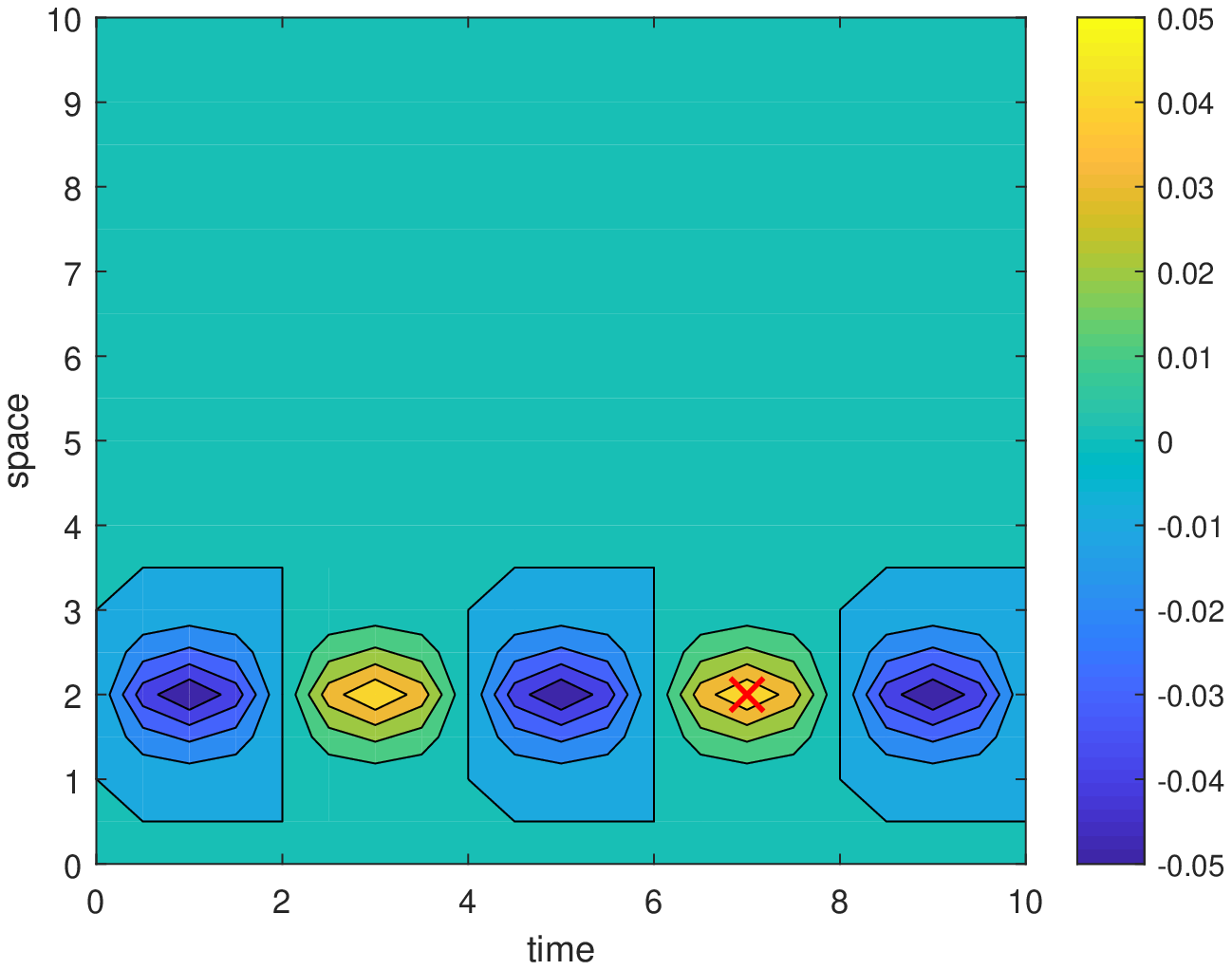}
  \caption{}
  \label{fig:covstr_c}
\end{subfigure}
\caption{Example illustrations of spatio-temporal covariance structures that are possible using the proposed function $\basis(\s,t)$. Contour plots of $\Cov_{\psub}[y,y']$ for $y$ at a test point $(s,t)$ (red cross) and $y'$ at all other coordinates $(s',t')$ in the space-time domain.
For sake of illustration, the spatial dimension is $d=1$. (a) Test point $(s,t)=(9,4)$. The test point is positively correlated with only neighbouring points in space and has a constant covariance with points across time. (b) Test point $(s,t)=(4.5,0)$. The covariance is local in space as in (a) but across time the covariance decays slowly. (c) Test point $(s,t)=(2,7)$. The covariance is periodic.}
\label{covstr}
\end{center}
\vskip -0.2in
\end{figure*}

Below we specify the function $\basis(\s,t)$ in \eqref{eq:modelclass} such that the spatio-temporal covariance structure $\Cov_{\psub}[y,y']$ can express local spatial patterns with varying temporal periodicities as illustrated in Figure~\ref{covstr}. Subsequently, we apply a covariance-fitting methodology for learning the model parameters such that the predictor \eqref{eq:predictor} can be updated sequentially for each new observation \cite{zachariah2017online}.

\subsection{Local-periodic space-time basis} \label{sptiotemporal basis}

The function $\basis(\s,t)$ varies over a space-time domain $\mathcal{S} \times \mathcal{T} \subset \mathbb{R}^{d+1}$ and its elements can be thought of as basis functions. It is formulated as a Kronecker product of a time and space bases,
\begin{equation} \label{space time basis}
\basis(\s,t)=\tbasis(t)\otimes\sbasis(\s),
\end{equation}
for compactness.

\begin{figure*}[ht!]
%\vskip 0.1in
\begin{center}
\begin{subfigure}{0.33\textwidth}
  \centering
  \includegraphics[width=0.75\linewidth]{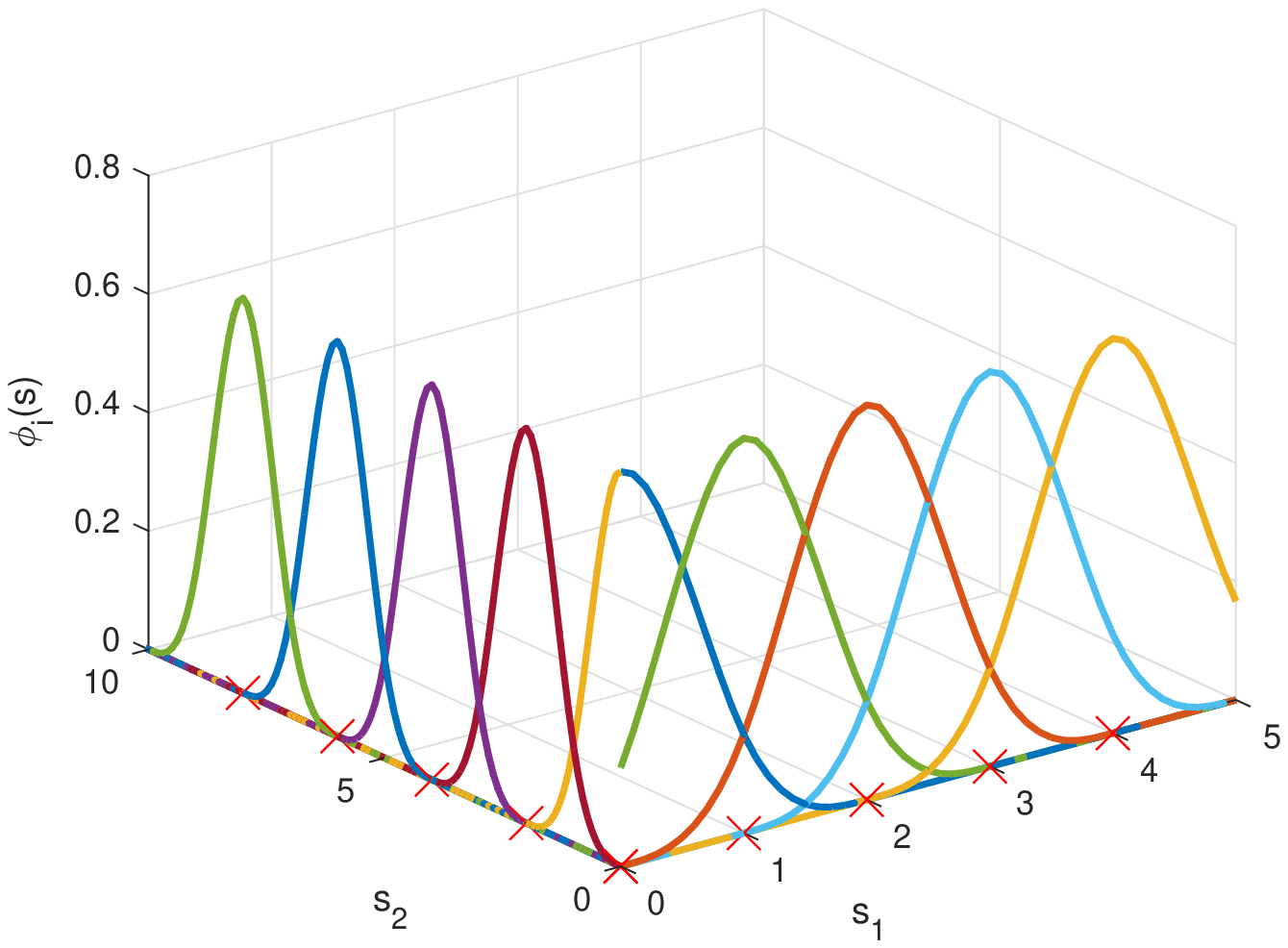}
  \caption{}
  \label{fig:bspline_2d}
\end{subfigure}
\begin{subfigure}{0.33\textwidth}
  \centering
  \includegraphics[width=0.75\linewidth]{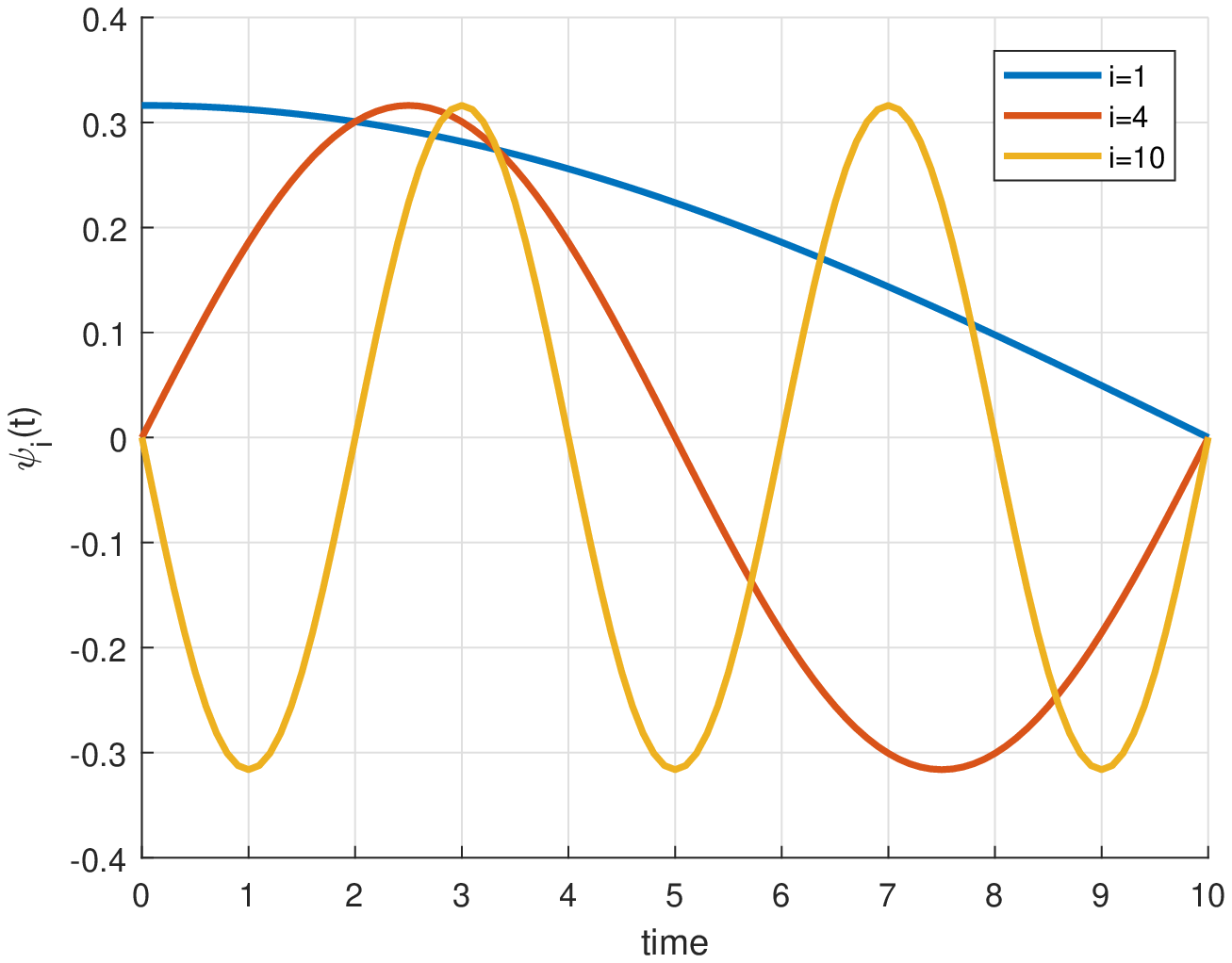}
  \caption{}
  \label{fig:laplace}
\end{subfigure}
\caption{(a) Components of the local spatial basis $\sbasis_1(s_1)$ and $\sbasis_2(s_2)$, respectively. Each component is centered at red crosses on the spatial axes $s_1$ and $s_2$. (b) Components of the periodic temporal basis $\tbasis(t)$.}
\label{basis}
\end{center}
\vskip -0.2in
\end{figure*}

We begin by specifying the spatial function as \begin{equation} 
\sbasis(\s) = \sbasis_1(s_1) \otimes \cdots \otimes \sbasis_d(s_d)
\end{equation}
where the basis vector for the $i^\text{th}$ spatial dimension,
\begin{equation}
    \sbasis_i(s_i) =  \col\{ \: \varphi_{i,1}(s_i) , \: \cdots , \: \varphi_{i,N_s}(s_i) \: \}
\label{eq:bsplinebasis}
\end{equation}

is composed of $N_s$ localized components with a finite support $L$. For notational simplicity, we consider $N_s$ and $L$ to be same for each dimension $i$. Based on their computational attractiveness and local approximation properties we use a cubic spline basis \cite{rasmussen2006gaussian,wasserman2006all}. Then \eqref{eq:bsplinebasis} is given by \eqref{eq:bspline}, where $c$ determines the location of a component. Figure~\ref{fig:bspline_2d} illustrates the components as a function of its spatial dimension. We place the centers $c$ of each component uniformly across the spatial dimensions. 

Using $\sbasis(\s)$ allows for covariance structures that are localized in space in such a way that neighbouring points have a nonnegative correlation and points far from each other have no correlation as determined by the support size $L$. Hence for a given $L$, the resulting covariance structure can capture local spatial patterns of a certain scale and can easily be extended to cover multiple scales by replacing \eqref{space time basis} with for example
  \begin{equation*}
    \basis(\s,t) = \tbasis(t)\otimes
    \begin{bmatrix}
           \sbasis_{L_1}(\s) \\
           \sbasis_{L_2}(\s) \\
         \end{bmatrix}
  \end{equation*}  
that accommodates two different support sizes $L_1$ and $L_2$.  
The number of basis functions $N_s$ is chosen such that adjacent localized components $\varphi_i(s)$ have overlapping support to cater for points in between them. This requirement is fulfilled by choosing $N_s>\frac{R_s}{L}$ where $R_s$ is the range of the spatial dimension. For example when $N_s=\frac{2R_s}{L}$, the adjacent component $\varphi_i(s)$ have $50$ percent overlap. The maximum value of $N_s$ is limited by the number of training points and the computational resources that are available. 

The temporal function $\tbasis(t)$ is also specified by a basis
\begin{equation}
    \tbasis(t) = \col\{\: \psi_0(t), \psi_1(t), \: \dots, \: \psi_{N_t}(t) \: \}.
\end{equation}
However, to be able to predict missing data of the type illustrated in Figure~\ref{fig:problemsetup} we cannot rely on a localized basis for extended interpolations over space-time. Due to its good approximating properties we instead apply the periodic basis developed by \citet{solin2014hilbert} defined over a range $\mathcal{T} = [0,R_t]$:
\begin{equation} \label{temporal basis}
  \tbasissc_k(t)=
  \begin{cases}
    1 & \text{$k=0$}\\
    \frac{1}{\sqrt{R_t}}\sin{(k\pi\frac{t+R_t}{2R_t})} & \text{otherwise}
  \end{cases}
\end{equation}
Similar to a Fourier basis, $\tbasis(t)$ allows for periodic covariance structures that capture both fixed and periodic patterns in the data along time with different frequencies. Moreover, as $N_t$ grows, any temporally stationary covariance structure can be captured, cf. \cite{solin2014hilbert}. Using \eqref{temporal basis}, the maximum frequency in the model is $\frac{N_t}{4R_t}$. Hence, depending on the data and the highest frequency periodic patterns we may expect in it, an appropriate value of $N_t$ can be chosen.

In summary, the proposed spatio-temporal basis $\basis(\s,t)$ in \eqref{space time basis} is of dimension $p = N^d_s (N_t+1)$ and yields a covariance function $\Cov_{\psub}[y,y']$ that may vary temporally with different frequencies specific to different spatial regions, as illustrated in Figure \ref{covstr}. The covariance structure is determined by the parameter $\pvec$, which we learn using a covariance-fitting methodology considered next.

\subsection{Learning method for streaming data} \label{online learning method}

We describe a covariance-fitting approach for learning the model parameter $\pvec$, up to an arbitrary scale factor, from streaming data. Given a training dataset $\dataset_n$, this approach enables us to update the predictor $\what{y}_{\psub}(\s, t)=\lin^\T_\psub(\s,t) \mbf{y}$ from \eqref{eq:predictor} in a streaming fashion as $n=1, 2, \dots$. We consider fitting the model covariance structure of the training data $\y$, which is parameterized by $\pvec$ in \eqref{eq:covmat}, to the empirical structure. Let us first define a normalized sample covariance matrix of the training data,
\begin{equation*}
    \samplecovmat=\frac{(\y-\1\eta)(\y-\1\eta)^\T}{\|\y-\1\eta\|_2}.
\end{equation*}
Here $\1$ corresponds to using $\mbf{u}(\s,t) \equiv 1$. Then the optimal model parameters are given by a covariance-fitting criterion (cf. 
\cite{cressie1985fitting, Anderson1989_linear, cressie2008fixed, StoicaEtAl2011_newspectral}) with minimizer:
\begin{equation} \label{objective function}
    \what{\pvec} \: =\: \argminA_{\pvec} \; \big\|\samplecovmat-\covmat_{\psub}\big\|^2_{\covmat^{-1}_{\psub}}
\end{equation}
Here the matrix norm corresponds to a weighted norm which penalizes correlated residuals. The learned parameter $\what{\pvec}$ is invariant with respect to the mean parameter $\eta$ and can be rescaled by an arbitrary scale factor  \cite{zachariah2017online}. Moreover, the resulting predictor corresponding to $\what{\pvec}$ in equation (\ref{eq:predictor}) can be written in the equivalent form:
\begin{equation}
\what{y}_{\what{\psub}}(\s,t) \; \equiv \; \mbs{\alpha}^\T(\s,t)\mbf{w}^\star 
\label{eq:predictor_equiv}
\end{equation}
where $\mbs{\alpha}(\s,t) = \col\{  1,\basis(\s,t)\}$. The $(p+1)$-dimensional weight vector $\mbf{w}^\star$ is defined as the minimizer
\begin{equation} 
    \mbf{w}^\star = \argminA_\mbf{w} \sqrt{\what{\E}\big[ |y_i - \mbs{\alpha}^\T(\s_i,t_i) \mbf{w}|^2 \big]} + \frac{1}{\sqrt{N}}\|\mbs{\zeta}\odot\mbf{w}\|_1
\end{equation}
where the elements of $\mbs{\zeta}$ are given by
\begin{equation*} 
  \zeta_j=
  \begin{cases}
    \frac{1}{\sqrt{N}}\|[\basismat]_{j-1}\|_2, & \text{$j>1$}\\
    0, & \text{otherwise}
  \end{cases}
\end{equation*}
For proofs of these relations and a derivation of its computational properties,see \citet{zachariah2017online}

The resulting predictor in \eqref{eq:predictor_equiv} is called the \SPICE{} (sparse iterative covariance-based estimation) predictor. It is computed via a convex and sparsifying regularized minimization problem that can be solved using coordinate descent with recursively updated quantities at each new training point  $(\s_n,t_n, y_n)$. By exploiting this structure, our predictor $\what{y}_{\what{\psub}}(\s,t)$ can now be updated with streaming data as $n=1, 2, \dots$. A pseudocode implementation is provided in Algorithm \ref{algorithm 1}. The key recursively updated quantities passed from one update to the next are the symmetric matrix $\mbf{\Gamma}$ and the vectors $\mbs{\rho}$ and $\check{\mbf{w}}$ of dimension $p+1$ along with the scalar $\kappa$. Here $\check{\mbf{w}}$ is the weight vector at sample $n-1$, which is initialized at zero along with the above variables in Algorithm \ref{algorithm 1}. The runtime is linear in $n$ and constant in memory. That is, for a fixed training data size $N$, the total runtime of the algorithm is on the order $\mathcal{O}(Np^2)$ and its memory requirement is $\mathcal{O}(p^2)$. For further details, we refer the reader to the supplementary material. Code available at \href{https://github.com/Muhammad-Osama/Localized-Spatio-temporal-Models}{github}.  

\begin{algorithm}
\begin{algorithmic}
\STATE Input: $(\s_n,t_n, y_n)$ and $\check{\mbf{w}}$
\STATE $\mbf{\Gamma}:=\mbf{\Gamma}+\mbs{\alpha}(\s_n,t_n)\mbs{\alpha}^\T(\s_n,t_n)$
\STATE $\mbs{\rho}:= \mbs{\rho}+\mbs{\alpha}(\s_n,t_n) y_n$
\STATE $\kappa:= \kappa+y_n^2$
\STATE $\epsilon:= \kappa+\check{\mbf{w}}^\T\mbf{\Gamma}\check{\mbf{w}}-2\check{\mbf{w}}^\T\mbs{\rho}$
\STATE $\mbs{\tau}:= \mbs{\rho}-\mbf{\Gamma}\check{\mbf{w}}$
\REPEAT
\STATE $j=1,\ldots,p+1$
\STATE $c_j:=\tau_j+\Gamma_{jj}\check{w}_j$
\IF{$j=1$}
\STATE $w_j :=\frac{c_j}{\Gamma_{jj}}$
\ELSE
\STATE $a_j:=\epsilon+\Gamma_{jj}\check{w}_j^2+2\check{w}_j\tau_j$
\STATE $\hat{s}_j:=\text{sign}(c_j)$
\STATE $\hat{r}_j:=\frac{|c_j|}{\Gamma_{jj}}-\frac{1}{\Gamma_{jj}}\sqrt{\frac{a_j\Gamma_{jj}-|c_j|^2}{n-1}}$
\STATE $w_j :=
  \begin{cases}
    \hat{s}_j\hat{r}_j & \text{$\sqrt{n-1}|c_j| > \sqrt{a_j\Gamma_{jj}-|c_j|^2}$}\\
    0 & \text{otherwise}
  \end{cases}$
%\end{equation*}
\ENDIF
\STATE $\epsilon:= \epsilon+\Gamma_{jj}(\check{w}_j-w_j^\star)^2+2(\check{w}_j-w_j^\star)\tau_j$
\STATE $\mbs{\tau}:= \mbs{\tau}+[\mbf{\Gamma}]_j(\check{w}_j-w_j^\star)$
%\STATE $\check{w}_j:=w_j^\star$
\UNTIL number of iterations equal L
\STATE Output: $\mbf{w}^\star = \check{\mbf{w}}$
\end{algorithmic}
\label{alg:spice}
\caption{Learning from streaming datasets}\label{algorithm 1}
\end{algorithm}

\section{Synthetic data}
\label{sec:synthetic}

The proposed method has been derived for predictions using large and/or streaming data sets. We now demonstrate its predictive properties using synthetic data and for the sake of reference compare it with a \GPR{} (Gaussian process regression) method using different covariance functions $\Cov[y,y']$.

\subsection{Damped planar wave}

\begin{figure*}[ht!]
\vskip 0.2in
\begin{center}
\begin{subfigure}{0.24\textwidth}
  \centering
  \includegraphics[width=0.92\linewidth]{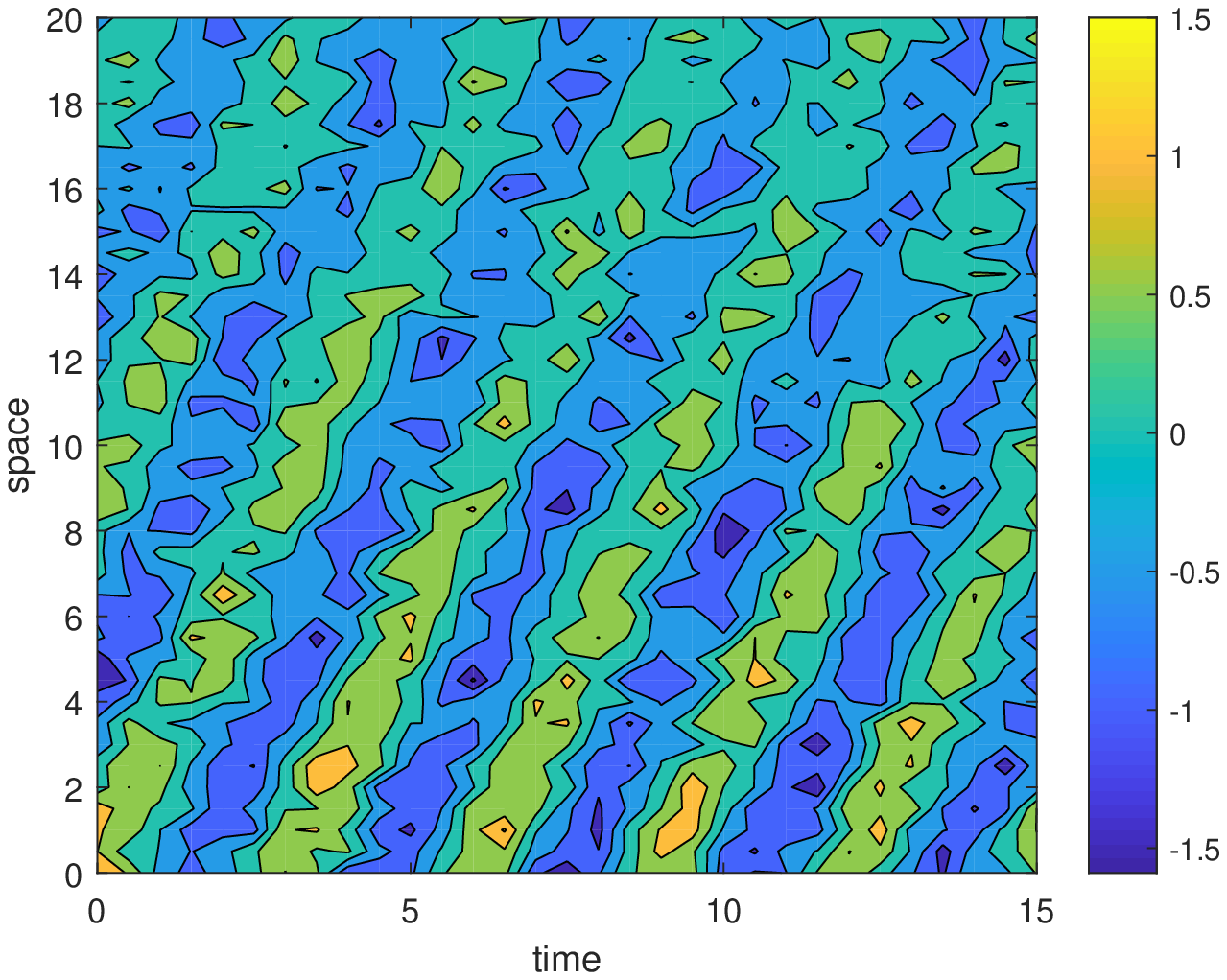}
  \caption{}
  \label{damping planar wave a}
 \end{subfigure} 
   \begin{subfigure}{0.24\textwidth}
  \centering
  \includegraphics[width=0.92\linewidth]{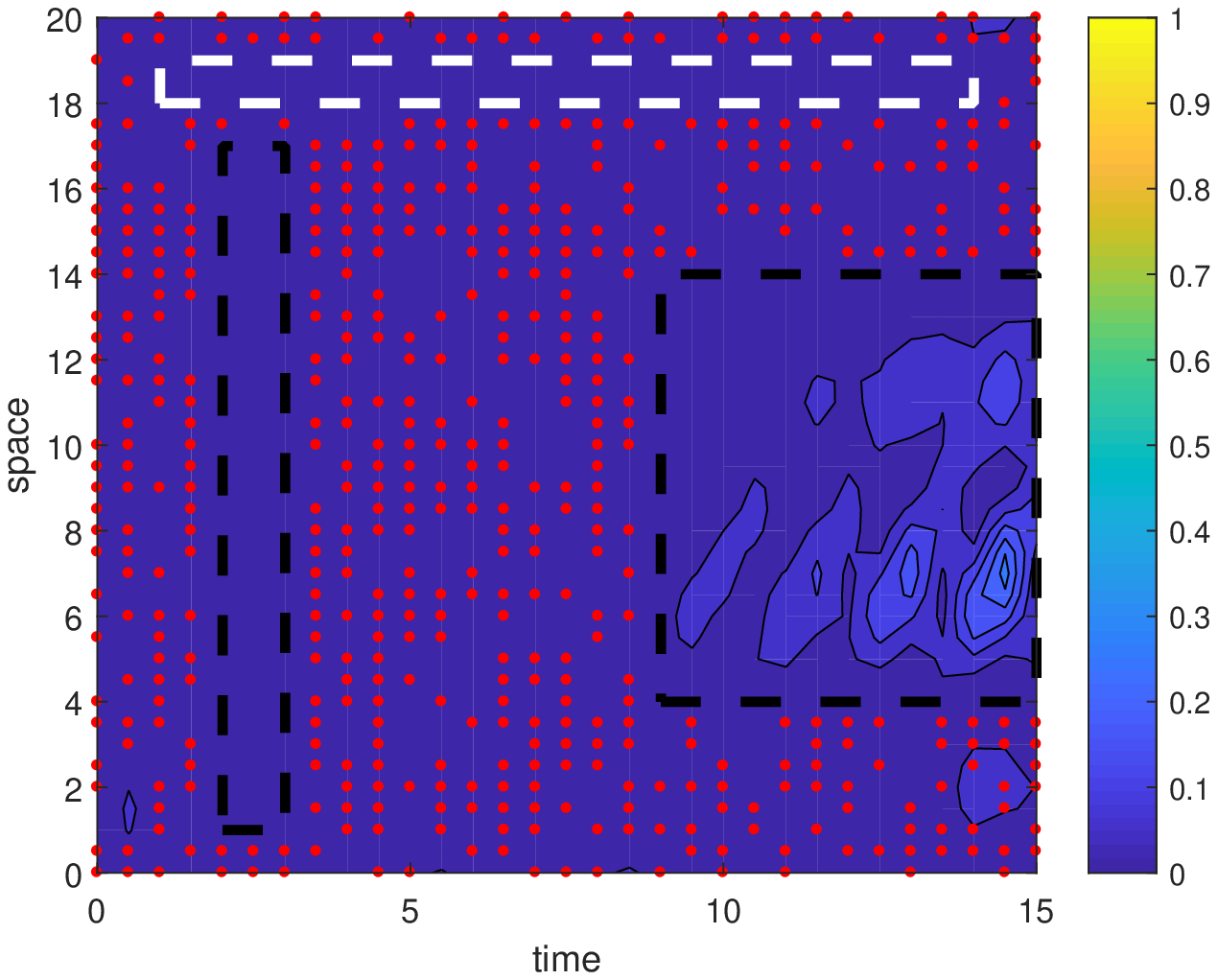}
  \caption{}
  \label{fig:damping planar wave_spice}
  \end{subfigure} 
   %\hfill
\begin{subfigure}{0.24\textwidth}
  \centering
  \includegraphics[width=0.92\linewidth]{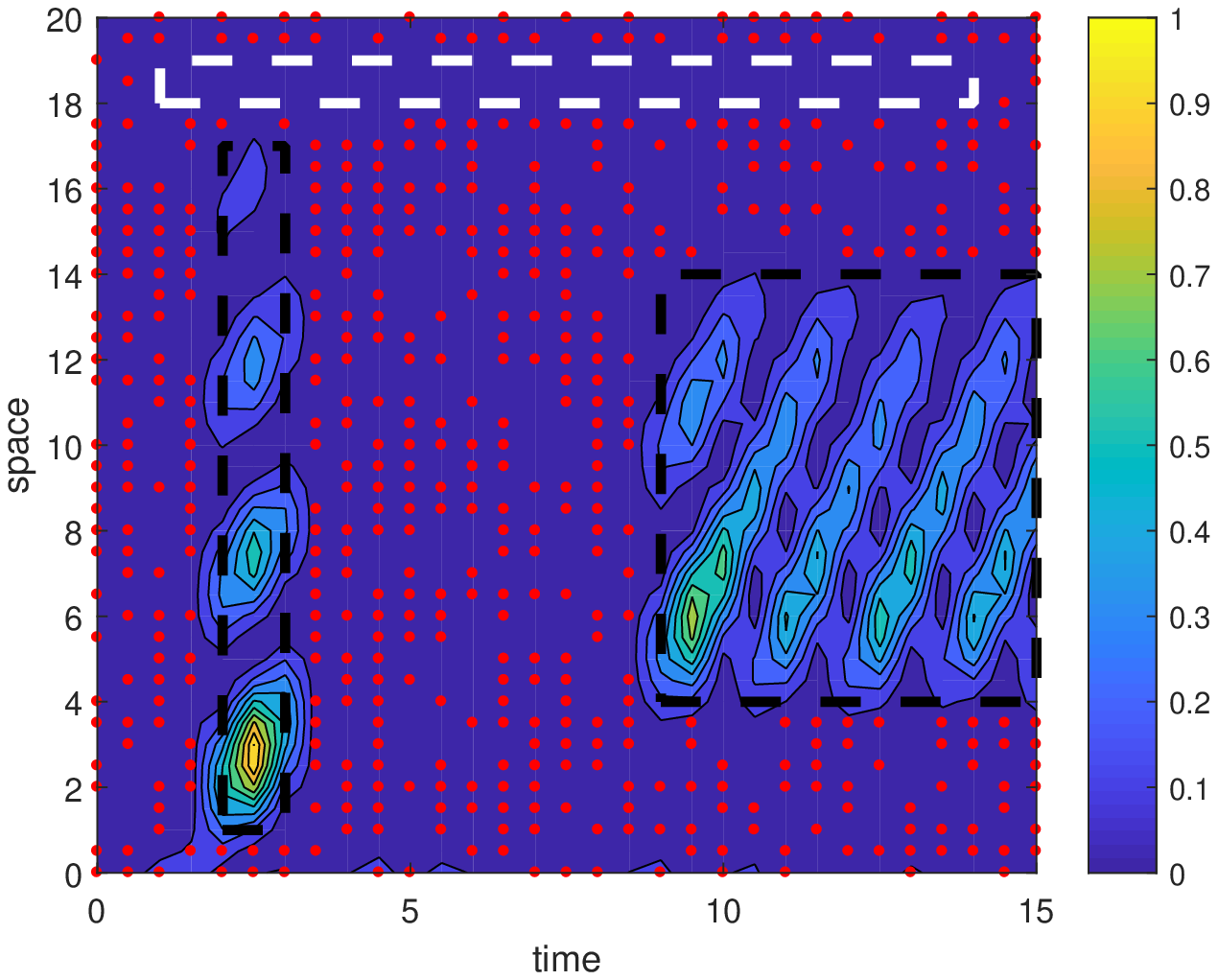}
  \caption{}
  \label{fig:damping planar wave_gpr}
  \end{subfigure} 
   %\hfill
  \begin{subfigure}{0.24\textwidth}
  \centering
  \includegraphics[width=0.92\linewidth]{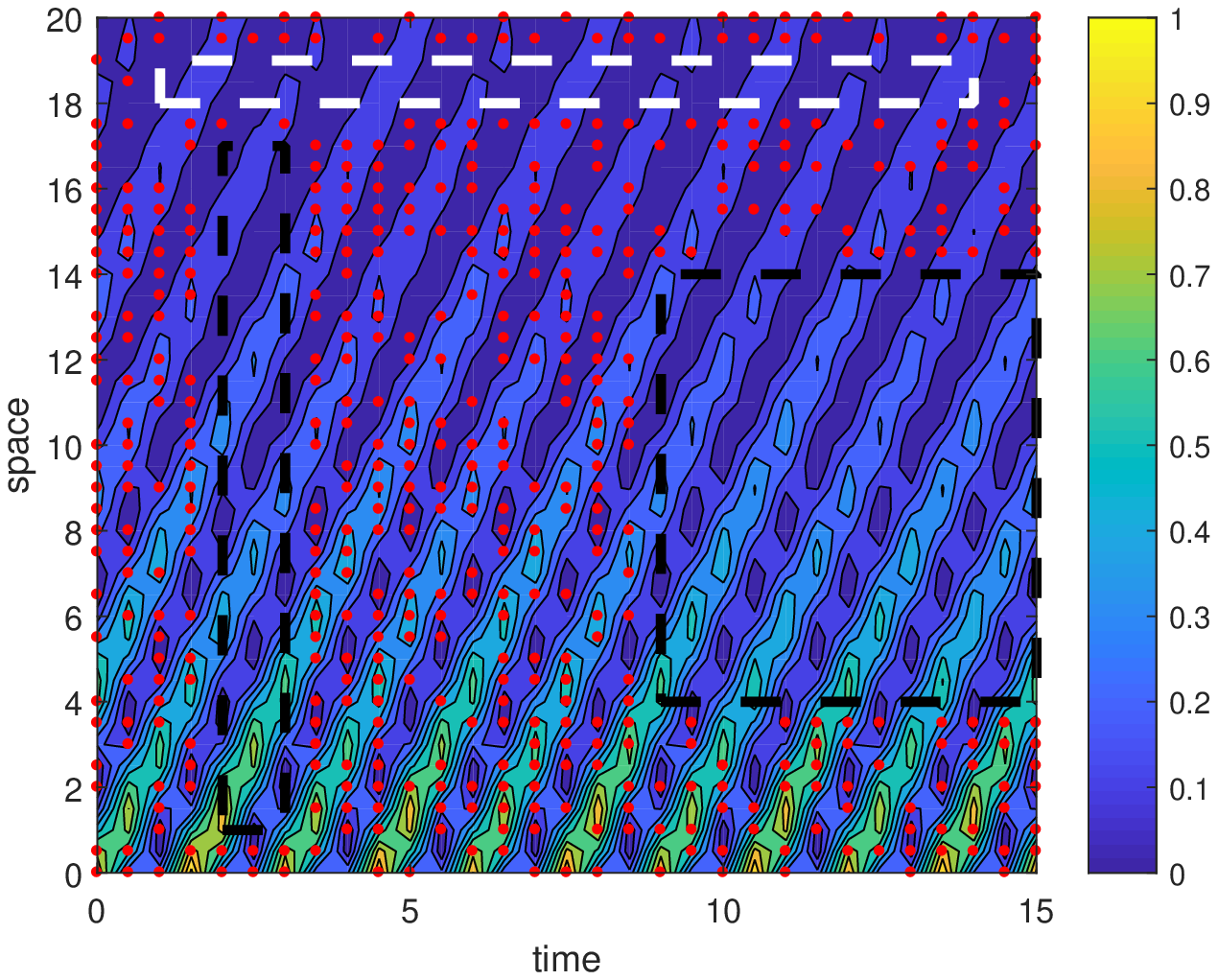}
  \caption{}
  \label{fig:damping planar wave_gpr_ard_period}
  \end{subfigure} 
  
\caption{(a) Realization of the damped planar wave process $y(s,t)$ defined in (\ref{eq:damping planar wave}). (b) MSE of the proposed method. The red dots denote training points.  The white and black dashed boxes represent contiguous space-time test regions where the data is missing. (c) MSE of \GPR{} using the Mat\'{e}rn ARD covariance model. (d) MSE of \GPR{} using using periodic Mat\'{e}rn ARD covariance function.}
\label{fig:damping planar wave}
\end{center}
\vskip -0.2in
\end{figure*}

To illustrate a dynamically evolving process, we consider planar a wave in one-dimensional space and time, cf. Figure~ \ref{damping planar wave a}. The unknown process is generated according to:
\begin{equation} \label{eq:damping planar wave}
    y(s,t) = \cos\left( \frac{2\pi}{\lambda_s}(s-v_st) \right) \exp\left(-\frac{s}{20}\right) + \varepsilon,
\end{equation}
where $v_s$ is the speed of the wave along space in units per second, $\lambda_s$ is the wavelength in units of space and $\varepsilon$ is a zero-mean white Gaussian process with standard deviation~$\sigma$. 

Note that the process decays exponentially as it propagates through space. For our experiments, we set $v_s = 3$~[spatial units/sec],~$\lambda_s=9$ [spatial units] and $\sigma~=~0.3$. Synthetic data is generated over a uniform grid and a subset of $N = 700$ training points are used. Different contiguous space-time blocks are selected as test regions to resemble realistic scenarios in which the coverage of sensors, satellites or other measurement equipment is incomplete. For example, the dashed white boxes in Figure~\ref{fig:damping planar wave} emulate cases where data over a small region is missing most of the time. By contrast, the dashed black boxes correspond to cases when data over large spatial regions is missing some of the time.

The process in these test regions as well as at other randomly missing points is predicted using the proposed method with $N_t=25$, $N_s = 15$ and a spatial basis support set to $L=5$ spatial units. This results in $\basis(s,t)$ being of dimension $p=N_s(N_t+1)=390$. The mean-square error (MSE) of the prediction is shown in Figure~\ref{fig:damping planar wave_spice} and evaluated using 25 Monte Carlo simulations. The region in the white box extends over almost the entire time dimension, hence there are very few neighbouring training points in time to draw upon for prediction and no information about the periodicities in the region. Instead our method leverages the neighbouring spatial information to obtain a good prediction resulting in a low MSE. Both black boxes are test regions that have neighbouring training points that provide temporal information about the process. However, left region has training points both before and after whereas the right region only has points before, yielding a more challenging prediction problem. Nevertheless, the proposed method is able to learn both the periodic and the local damping patterns to provide accurate predictions in both regions.

We include also the MSE of \GPR{} using two different covariance functions learned by a numerical maximum likelihood search. While this method is not applicable to the streaming data of interest here, it provides a performance reference. First, we use a Mat\'{e}rn ARD covariance model \cite{rasmussen2010gaussian} to carefully adapt both space and time dimensions.  In Figure~\ref{fig:damping planar wave_gpr} it is seen that the resulting prediction errors are markedly worse for the large missing spatial regions and the method naturally fails to capture the periodic pattern of the process. Next, we use a periodic Mat\'{e}rn ARD covariance model to also capture space-time periodicity. However, the MSE (Figure \ref{fig:damping planar wave_gpr_ard_period}) is degraded throughout, which is possibly due to the non-convex optimization problem used to learn the model parameters. It may lead to local minima issues, including learning erroneous periods.

\subsection{Varying seasonalities across space} 

\begin{figure*}[ht!]
\vskip 0.2in
\begin{center}
\begin{subfigure}{0.24\textwidth}
  \centering
  \includegraphics[width=0.92\linewidth]{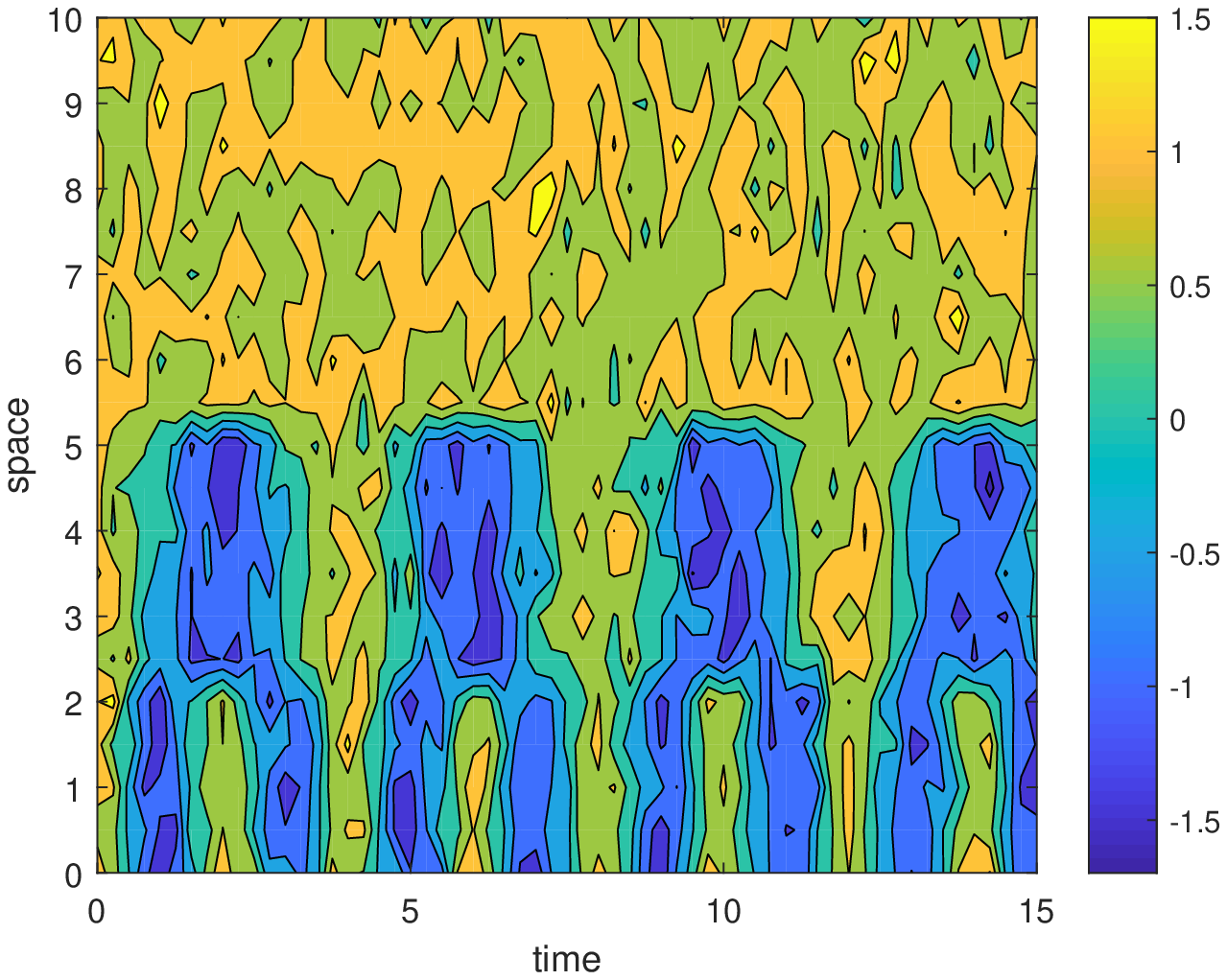}
  \caption{}
  \label{GP multiple periods a}
\end{subfigure}
\begin{subfigure}{0.24\textwidth}
  \centering
  \includegraphics[width=0.92\linewidth]{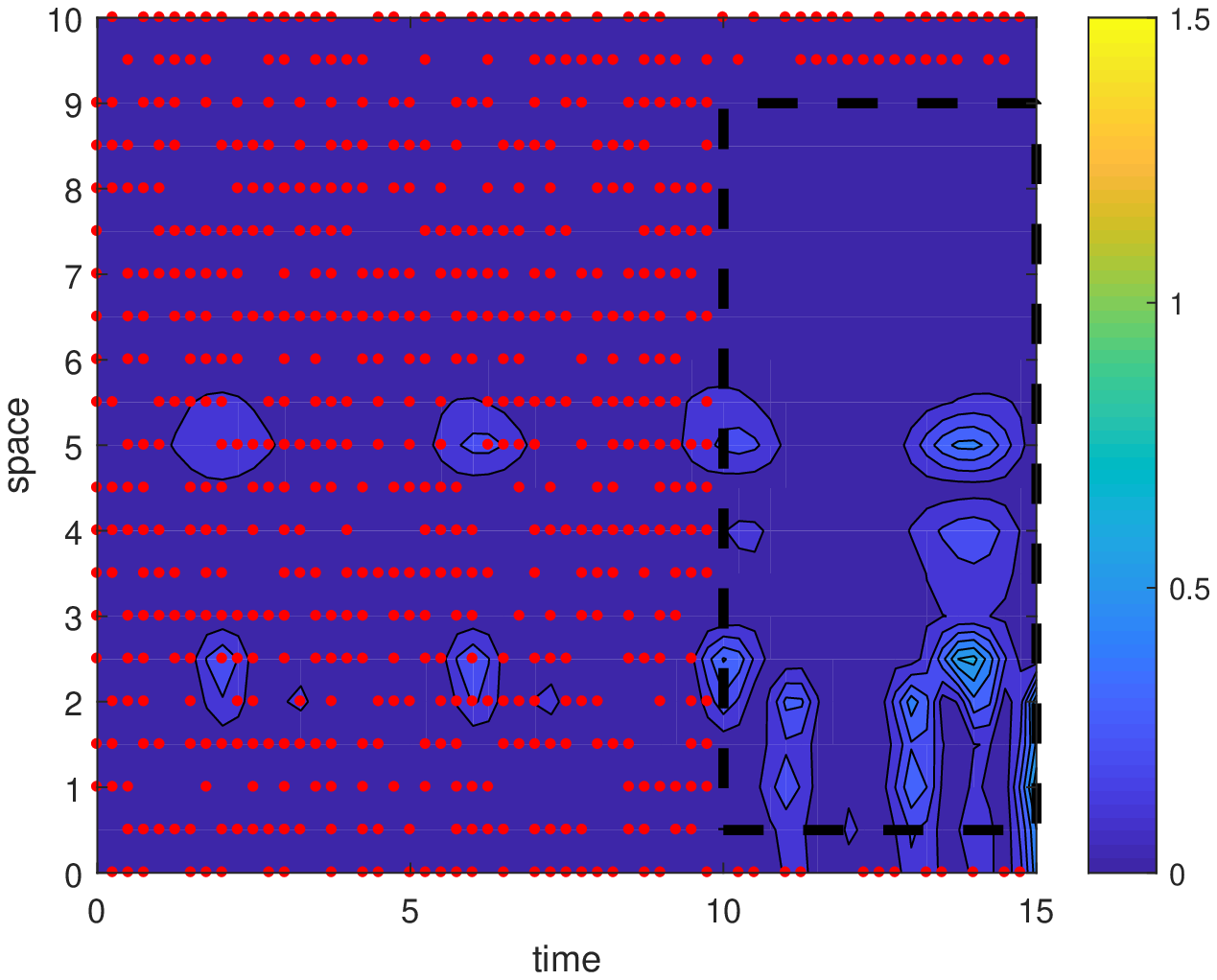}
  \caption{ }
  \label{fig:GP multiple periods_spice}
\end{subfigure}
%\hfill
\begin{subfigure}{0.24\textwidth}
  \centering
  \includegraphics[width=0.92\linewidth]{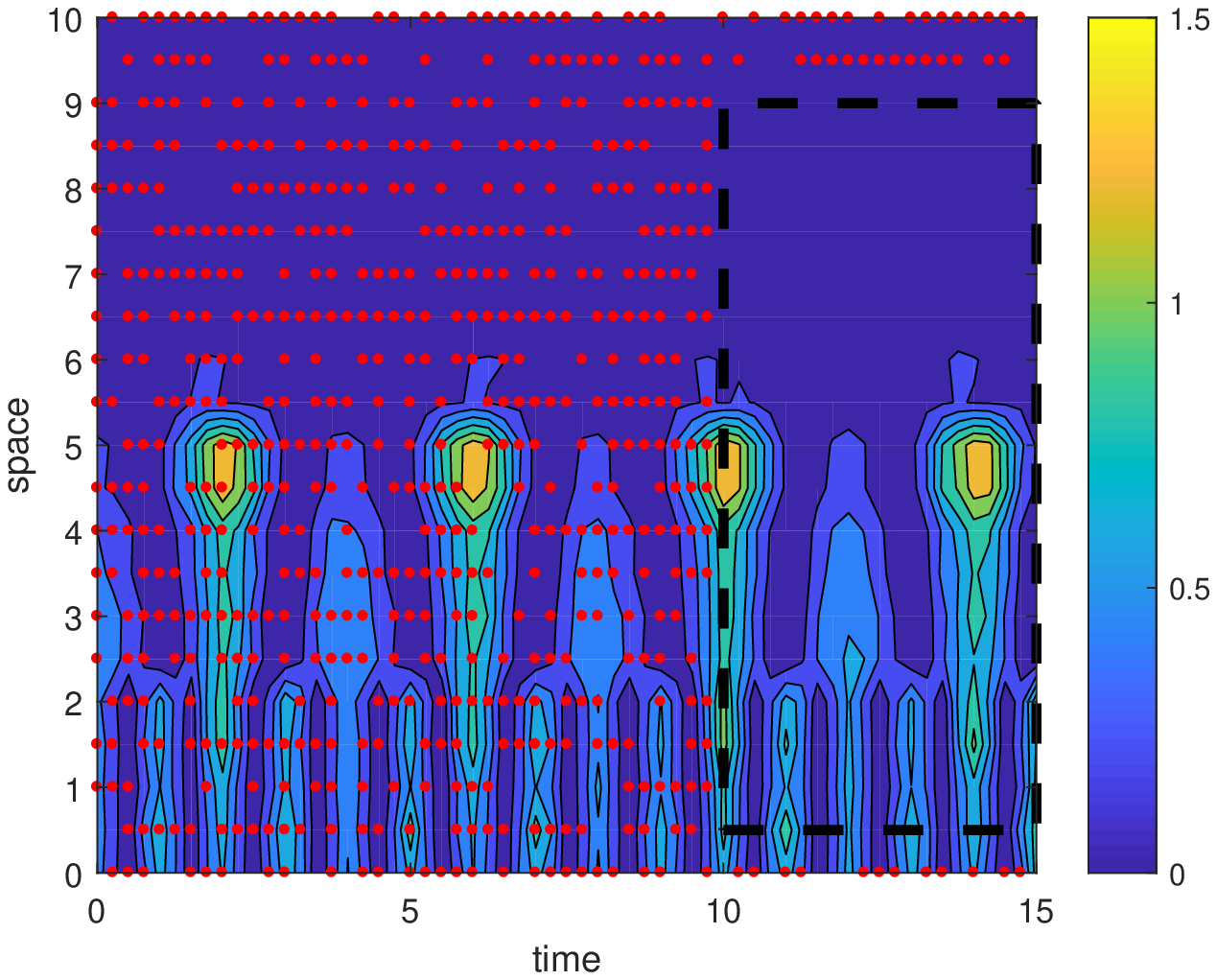}
  \caption{}
  \label{fig:GP multiple periods_gpr}
\end{subfigure}

\caption{(a) Realization of the process $y(s,t)$ defined in (\ref{eq:varying seasonalities across space}) with varying periods across space. (b) MSE of the proposed method which is able to learn different periodic temporal patterns across space. The red dots denote training points. The black dashed box marks a contiguous test region. (c) MSE of \GPR{} using periodic Mat\'{e}rn ARD covariance model.  }
\label{fig:GP multiple periods}
\end{center}
\vskip -0.2in
\end{figure*}

Here we generate a process that emulates scenarios of temporal periodicities which may vary across spatial regions. This occurs e.g. in climate data. Figure \ref{GP multiple periods a} shows a realization of a process generated according to 
\begin{equation} \label{eq:varying seasonalities across space}
    y(s,t) =\cos\left({\frac{2\pi}{T(s)}t}\right)+\varepsilon
\end{equation}
where the period $T(s)$ differs across space and $\varepsilon$ is zero-mean white Gaussian process with standard deviation $\sigma = 0.3$. In the upper region of the spatial domain $T(s) = \infty$, i.e., the process has a constant mean. In the middle and bottom regions $T(s)$ is large and small, respectively. The data is generated over a uniform grid and a subset of $N = 600$ points is used for training. A contiguous space-time block, marked by the dashed black box in Figure \ref{fig:GP multiple periods}, forms a test region to emulate scenarios where data can be missing over a large spatial region for some time.

For the proposed method we use $N_t=35$, $N_s=15$ and a support of $L=3$ for the spatial basis, so that $p=N_s(N_t+1)=540$. For the \GPR{} we use the periodic Mat\'{e}rn ARD kernel. Figures \ref{fig:GP multiple periods_spice} and \ref{fig:GP multiple periods_gpr} show the MSE performance of the proposed method and \GPR{} respectively which were obtained using 25 Monte Carlo simulations. The MSE of the proposed method is overall lower than that of \GPR{}, both in the dashed test region as well as outside it. Unlike the proposed method, \GPR{} has one parameter to fit to an overall periodic pattern and is thus unable to learn spatially localized patterns. Thus after learning, the process is predicted to be be nearly constant along time for all parts of the spatial region.

\section{Real data}
\label{sec:realdata}

We now demonstrate the proposed method for much larger, and possibly streaming, real-world datasets. 

\subsection{Pacific Sea Surface Temperature}

As a first application example, we use tropical pacific Sea Surface Temperature (SST) data \cite{SSTdata_source}. These data represent gridded monthly SST anomalies, in $^\circ$C, from January 1970 through March 2003 over a spatial region from $29^\circ$\mapS{} to $29^\circ$\mapN{} and $124^\circ$\mapE{} to $70^\circ$\mapW. The spatial resolution of the data is $2^\circ$ in both latitude and longitude.

\begin{figure}[h!]
%\vskip 0.2in
%\begin{center}
\begin{subfigure}{\columnwidth}
  \centering
  \includegraphics[width=0.58\columnwidth]{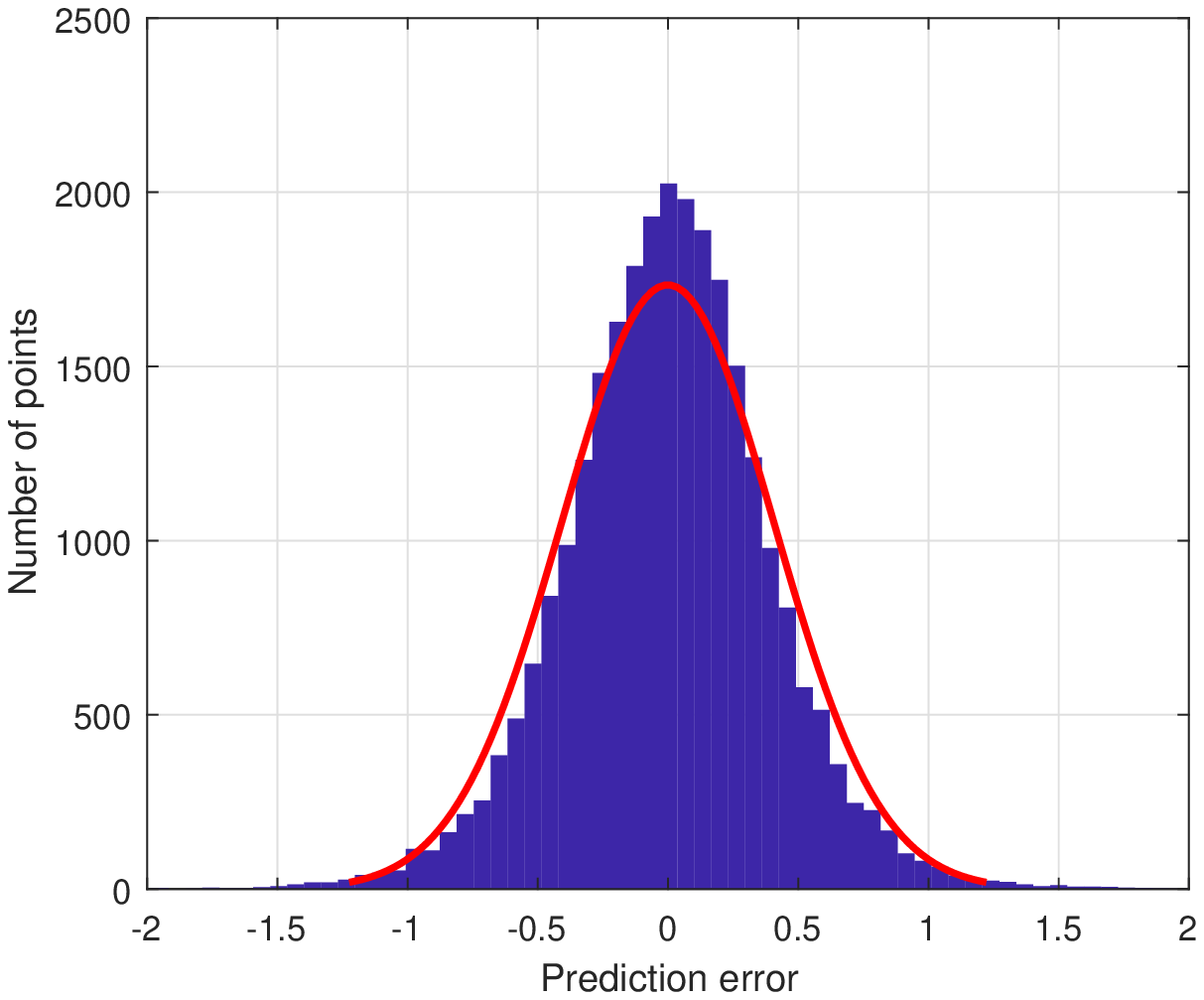}
  \caption{}
  \label{error-histogram-SST-simple}
  \end{subfigure}
  \hfill
\begin{subfigure}{\columnwidth}
  \centering
  \includegraphics[width=0.58\columnwidth]{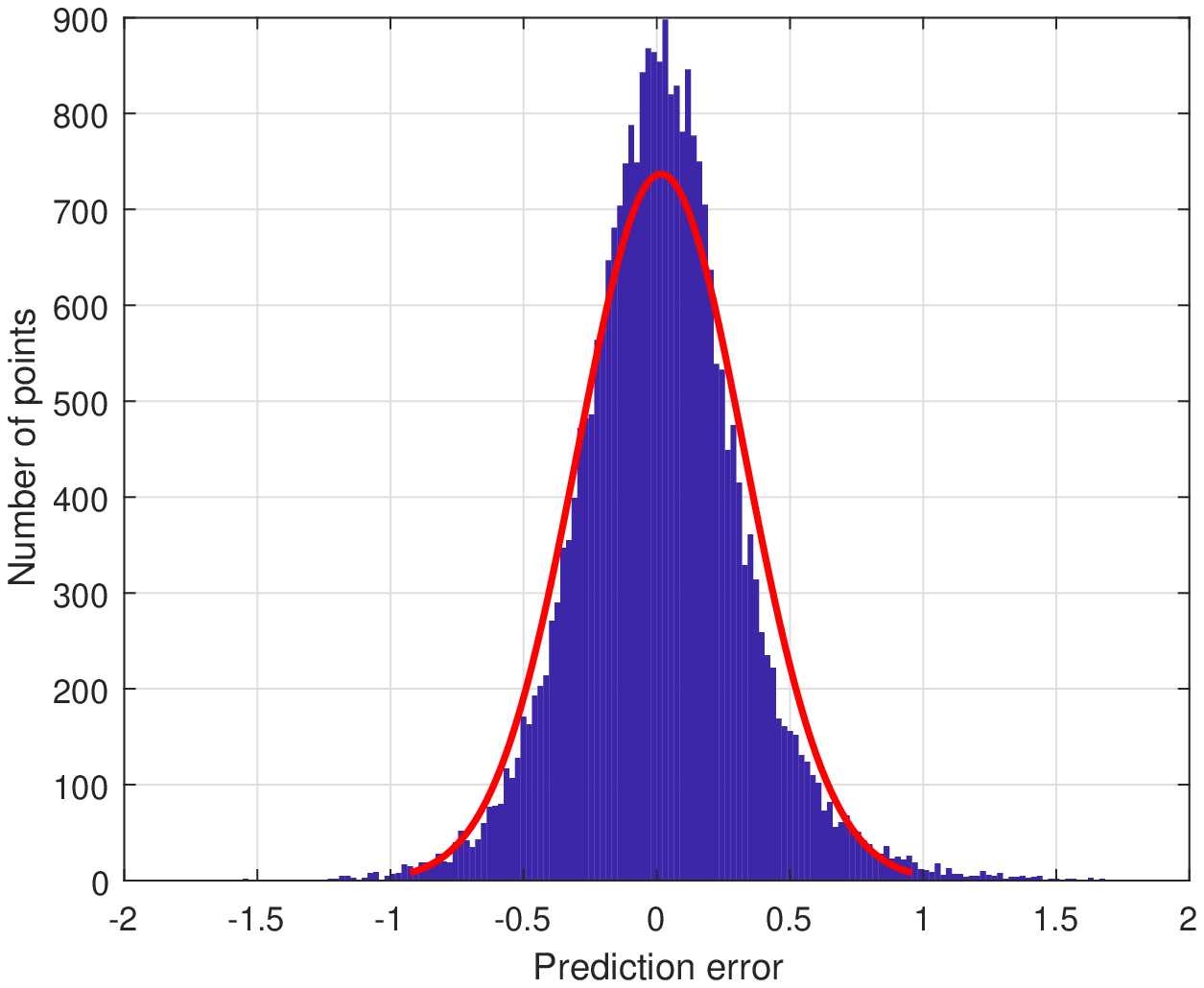}
  \caption{}
  \label{error-histogram-SST-spt-block}
  \end{subfigure}
  \hfill
\begin{subfigure}{\columnwidth}
  \centering
  \includegraphics[width=0.58\columnwidth]{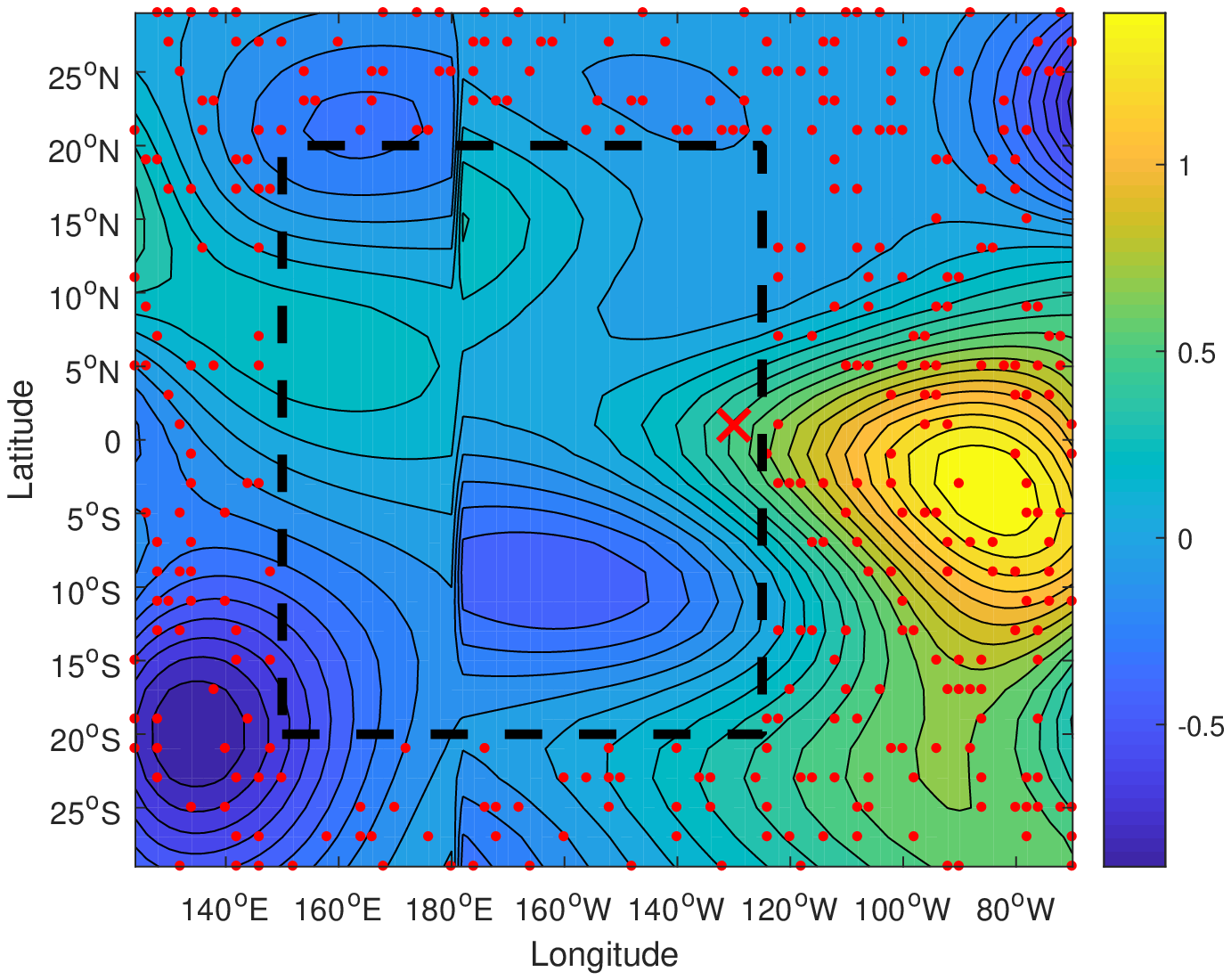}
  \caption{}
  \label{ypred-SST}
  \end{subfigure}
  \hfill
\begin{subfigure}{\columnwidth}
  \centering
  \includegraphics[width=0.58\columnwidth]{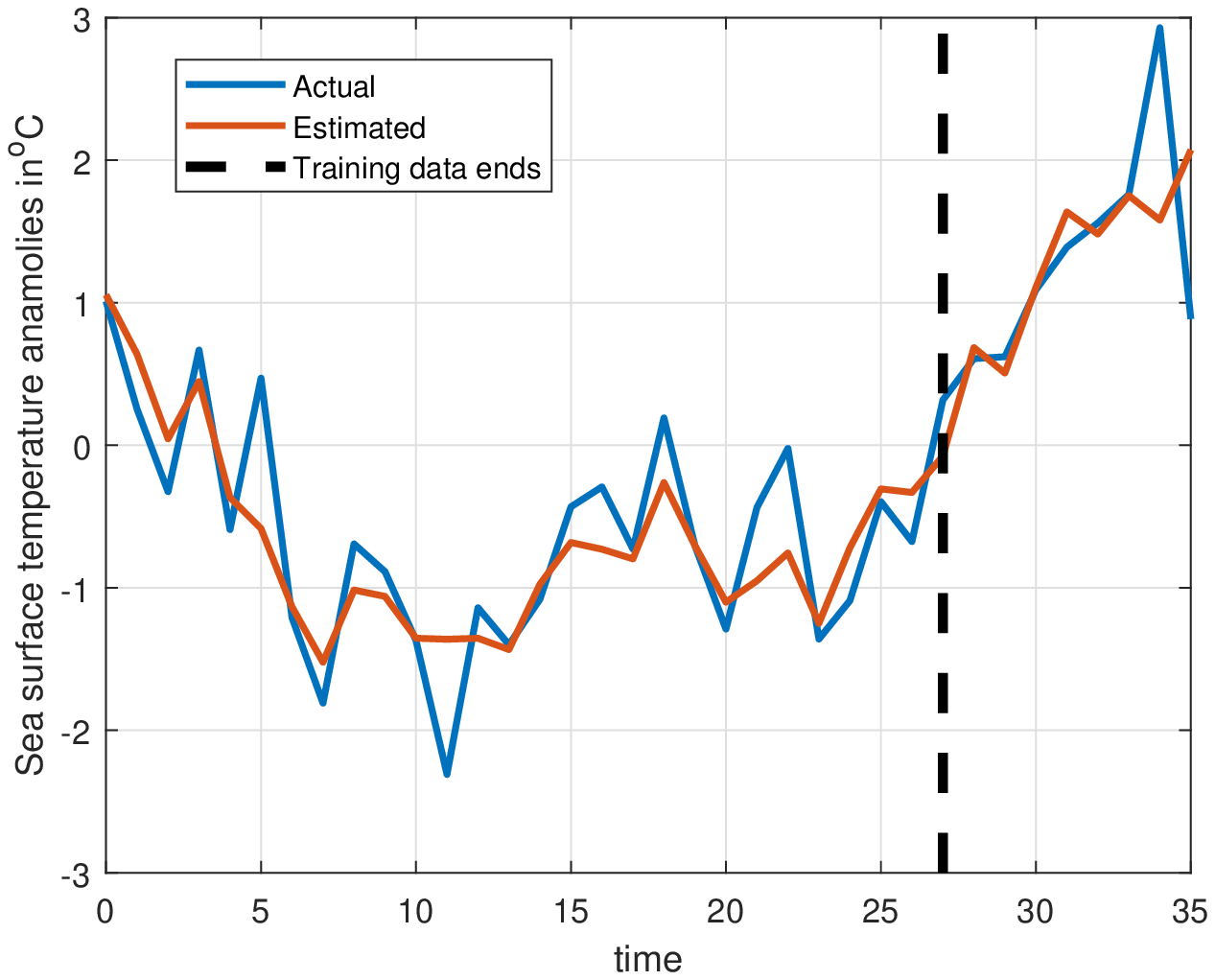}
  \caption{}
  \label{ypred-SST-time-series}
  \end{subfigure}
\caption{(a) Histogram of prediction error of all test points in the first scenario with randomly sampled training points. The plot in red is a fitted Gaussian distribution. Note that the dynamic range of the data is $[-3.2, 3.2]$ $^\circ$C . (b) Histogram of prediction error of all test points for the second scenario with a contiguous space-time block as test region. The plot in red is a fitted Gaussian distribution (c) Contour plot of predicted SST for a single spatial slice at time $t=30$. The red dots denote training points and the black dashed box indicates a contiguous test region. The red cross denotes a point of interest in which the El Ni\~{n}o effect can be observed. (d) Comparison of time series of actual and estimated SST for the point marked by the red cross in \ref{ypred-SST}. The data to the right of the black dashed line is part of the contiguous test region and is not used during training.}
\label{SST}
%\end{center}
%\vskip -0.2in
\end{figure}

Here we consider data from the first 36 months, making the total number of space-time data points equal to $36\times2~520=90~720$. In the first experiment, training points are sampled randomly across space-time and the missing data constitute the test points. Here we set $N~=~63~503$ as the number of training points. For the proposed method we set $N_t~=~100$, $N_s~=~8$ and the spatial support $L$ to be half of each spatial dimension. Then $p=N^2_s(N_t+1) = 6~464$. Figure~\ref{error-histogram-SST-simple} shows the prediction error histogram of all test points across the spatio-temporal domain. We see that it is centered around zero and its dispersion is considerably narrower than the dynamic range of the data. 

In the second experiment, we select a contiguous space-time block as a test region in addition to other test points to evaluate the performance in scenarios where data over entire spatial regions are missing for a period of time. Data falling within the spatial region marked by the black dashed box in Figure~\ref{ypred-SST} is missing beyond month 26, as indicated by the black dashed line in Figure \ref{ypred-SST-time-series}. Here $N=18~144$ are the number of training points. The prediction error histogram for this second experiment is shown in Figure~\ref{error-histogram-SST-spt-block} and remains fairly narrow. Figure \ref{ypred-SST} illustrates the predicted SST anomalies [$^\circ$C] for a spatial slice at month $t=30$. We pick a spatial point in a region where the El Ni\~{n}o effect, i.e., the periodic warming of the equatorial Pacific Sea Surface \cite{sarachik2010nino}, is known to be noticeable. The prediction of the SST anomalies at this spatial location across time along with the true SST is illustrated with Figure \ref{ypred-SST-time-series}. Note that the predictor is able to track the rising temperature deviation also for the missing data.

\subsection{Precipitation data}

As a second application example, we use precipitation data from the Climate Research Unit (CRU) time series datasets of climate variations \cite{jones2013cru}.  The precipitation data consists of monthly rainfall in millimeter over a period from $1901$ to $2012$ obtained with high spatial resolution (0.5 by 0.5 degree) over the whole planet. Here we consider a five year period from $2001$ to $2005$ and between spatial coordinates $95^\circ$\mapW{} to $107^\circ$\mapW{} and $40^{\circ}$\mapN~to $50^\circ$\mapN. This yields a total number of $28~800$ data points. 

\begin{figure}[h!]
%\vskip 0.2in
%\begin{center}
\begin{subfigure}{\columnwidth}
  \centering
  \includegraphics[width=0.58\columnwidth]{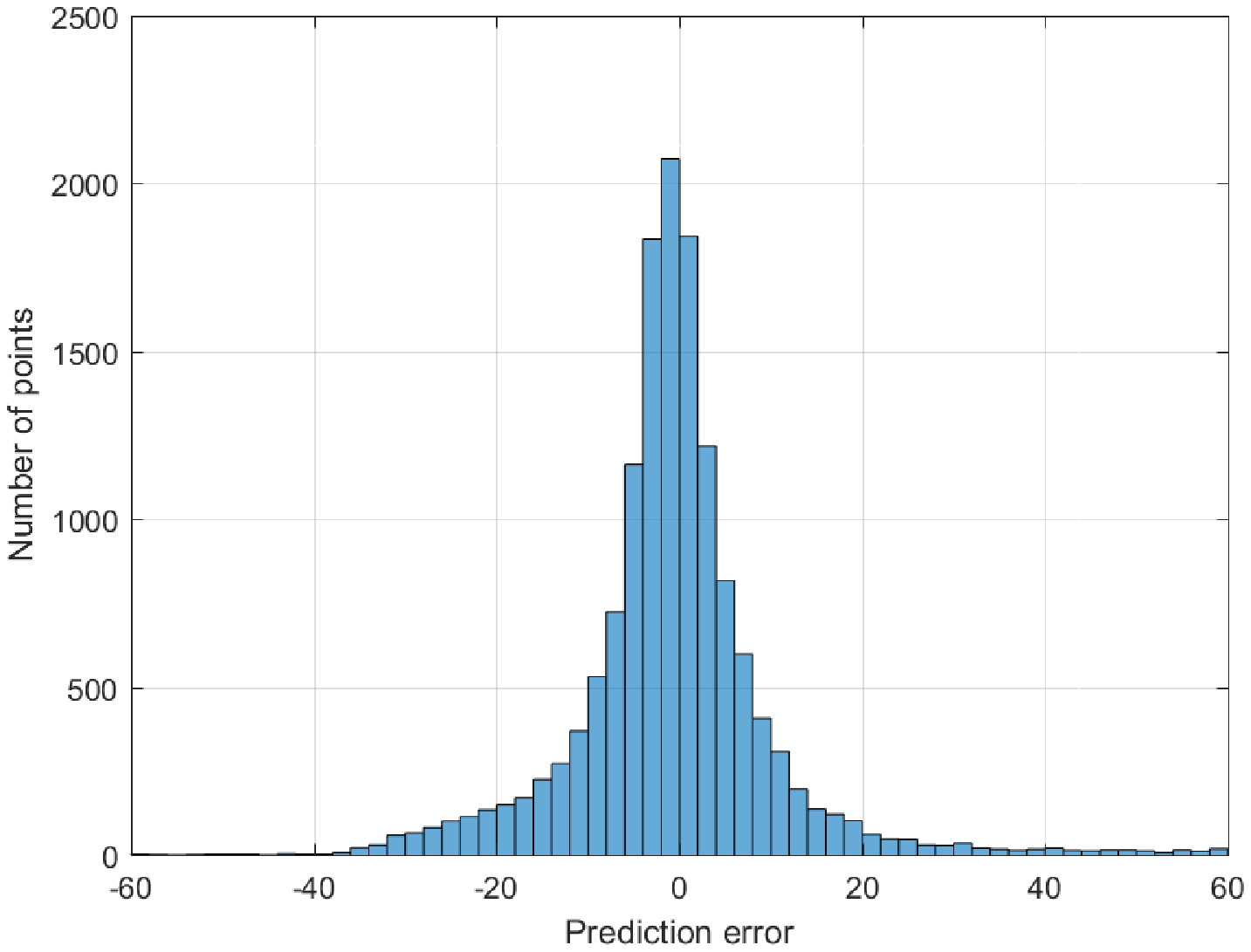}
  \caption{}
  \label{error-hist-pre}
  \end{subfigure}
  \hfill
\begin{subfigure}{\columnwidth}
  \centering
  \includegraphics[width=0.58\columnwidth]{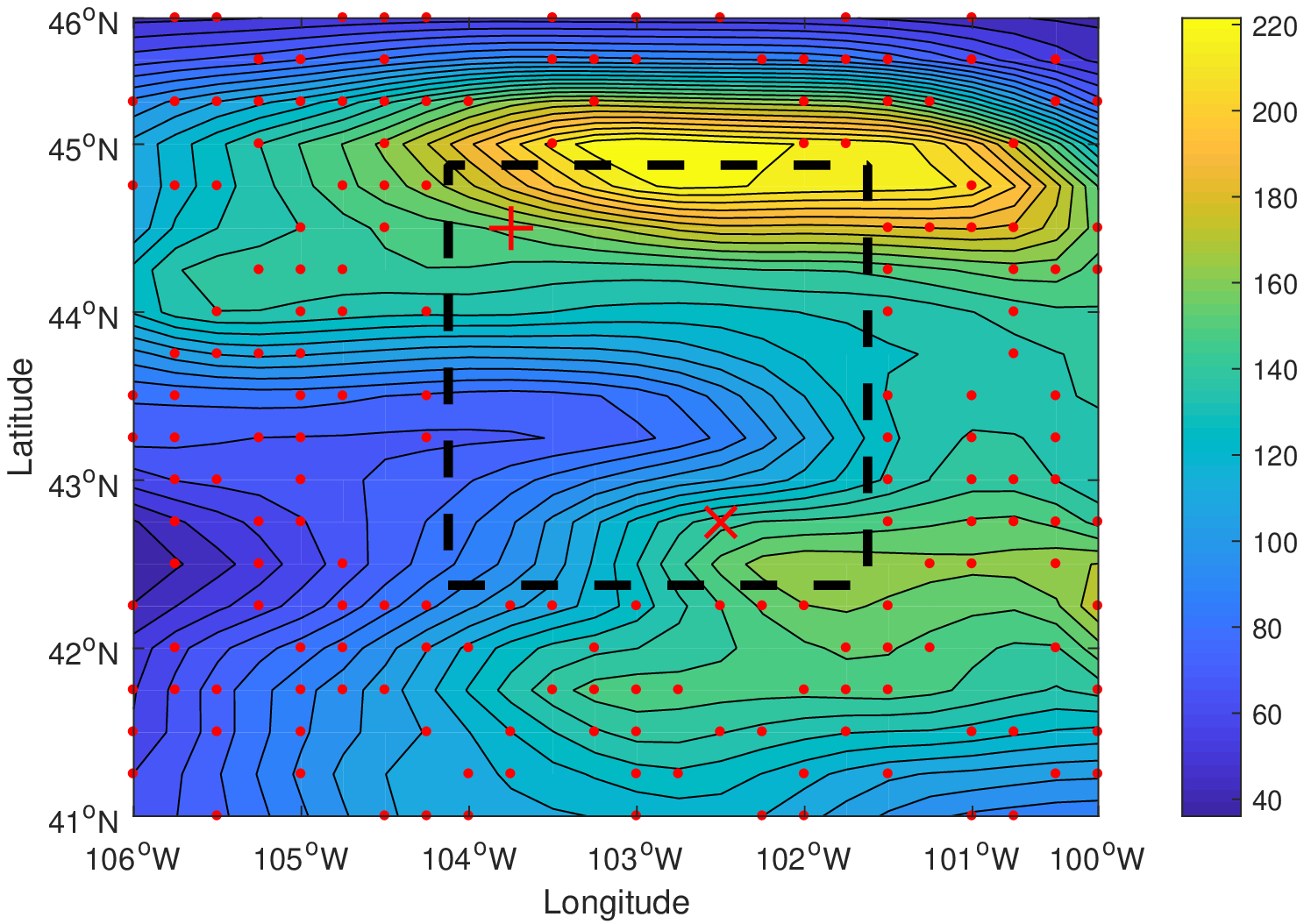}
  \caption{}
  \label{ypred-precipitaton}
  \end{subfigure}
   \hfill
\begin{subfigure}{\columnwidth}
  \centering
  \includegraphics[width=0.58\columnwidth]{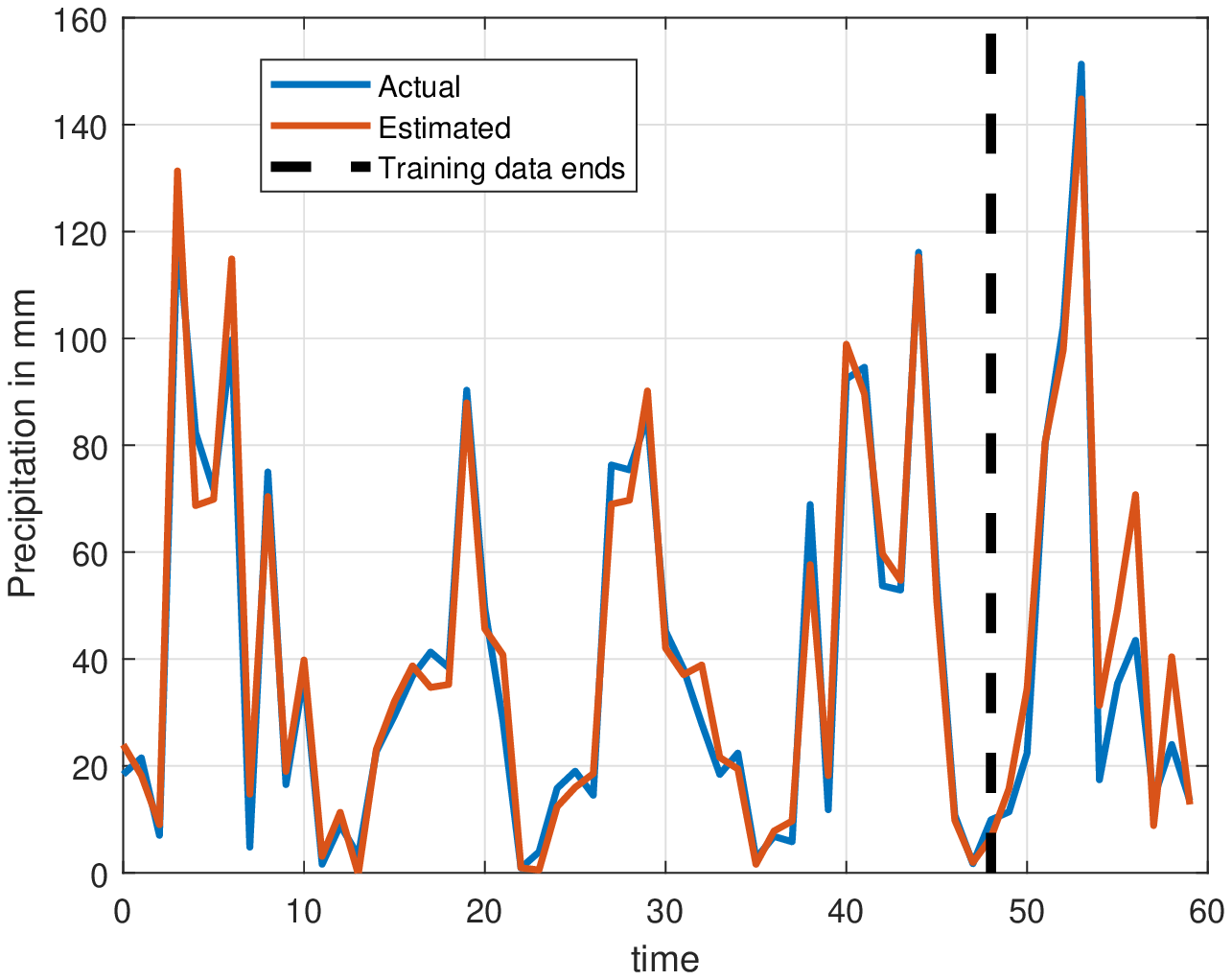}
  \caption{}
  \label{ypred-time-series-black-marker}
  \end{subfigure}
 \hfill
\begin{subfigure}{\columnwidth}
  \centering
  \includegraphics[width=0.58\columnwidth]{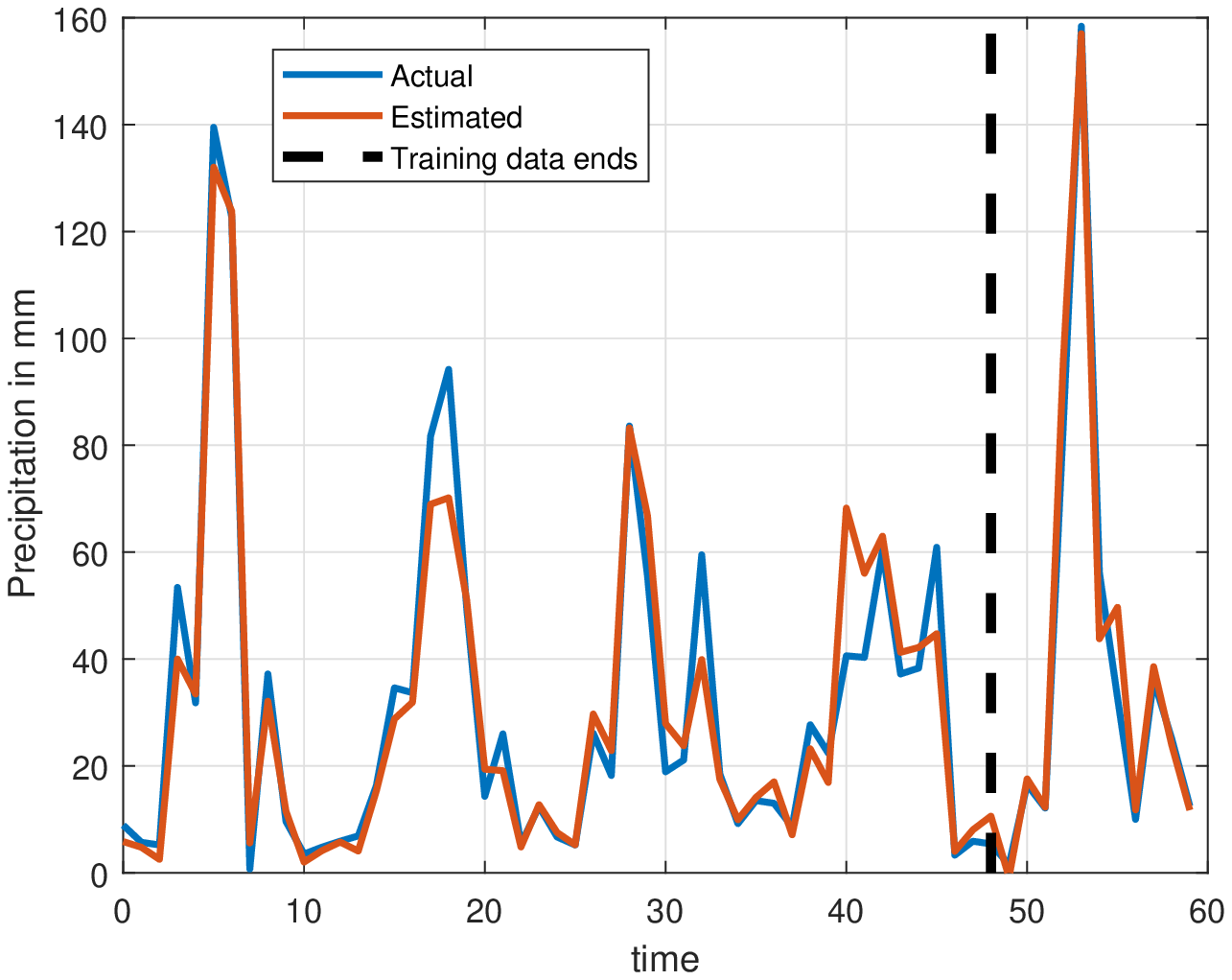}
  \caption{}
  \label{ypred-time-series-green-marker}
  \end{subfigure}
\caption{(a) Histogram of prediction error for all test points of precipitation data. The dynamic range of the data is $[0, 226]$ millimeters (b) Contour plot of predicted precipitation for a single spatial slice at time $t=54$. Data inside the black dashed box marks the contiguous test region and is not seen during training. The red dots denote training points. (c) The actual and estimated precipitation time series for the the spatial point marked by red cross in figure \ref{ypred-precipitaton}. (d) The actual and estimated precipitation time series for the the spatial point marked by red addition sign in figure \ref{ypred-precipitaton}. The data to the right of the black dashed line is part of the contiguous test region and is not used during training.}

\label{precipitation}
%\end{center}
%\vskip -0.2in
\end{figure}

The spatial region indicated by the black dashed box in Figure~\ref{ypred-precipitaton} beyond month $t=47$, as seen in Figure \ref{ypred-time-series-black-marker}, constitutes a contiguous test region, in addition to other randomly selected test points. The remaining $N = 14~400$ points are used for training.

For the proposed method we set $N_t~=~300$, $N_s~=~6$ and the spatial support $L$ to be half of each spatial dimension. Then $p=N^2_s(N_t+1) = 10~836$. Figure \ref{error-hist-pre} shows the prediction error histogram for the precipitation  test data. It is centered around zero and its dispersion is narrower than the dynamic range of the data. Figure \ref{ypred-precipitaton} shows the contour plot of predicted precipitation for a spatial slice at month $t=54$. The red cross and plus marker indicate spatial points whose actual and predicted time series are compared in Figures \ref{ypred-time-series-black-marker} and \ref{ypred-time-series-green-marker}, respectively. Note that the estimated precipitation tracks the true precipitation well everywhere even to the right of the black dashed line where the data was not seen during training. Note the ability of the predictor to track the different seasonal patterns in the missing regions.

\section{Conclusion}
We proposed a method in which a spatio-temporal predictor $\what{y}_{\what{\psub}}(\s, t)$ can be learned and updated sequentially as spatio-temporal data is obtained as a stream. It is capable of capturing spatially varying temporal patterns, using a non-stationary covariance model that is learned using a covariance-fitting approach. We demonstrated, using both simulated and real climate data, that it is capable of producing accurate predictions in large unobserved space-time test regions. In future work, we intend to further improve the computational efficiency of the method by exploiting the spatially localized structure of the covariance model.     

\newpage

%\newpage
\section*{Acknowledgements}
This research was financially supported by the project \emph{NewLEADS - New Directions in Learning Dynamical Systems} (contract number: 621-2016-06079), funded by the Swedish Research Council.

\nocite{langley00}
\clearpage
\bibliography{main_arxiv}
\bibliographystyle{icml2017}

\section*{Learning Localized Spatio-Temporal Models From Streaming Data: Supplementary Material}

\section{\textsc{Reformulating of fitting criterion}}

By expanding the objective function in (11), we obtain the equivalent form
\begin{equation*}
    \wtilde{\y}^\T\covmat^{-1}_{\psub}\wtilde{\y}+\underbrace{n\theta_0+\sum_{j=1}^{p}\theta_j\|[\basismat]_j\|_2}_{=\tr\{\covmat_\psub \}}
    \tag{16}
\end{equation*}
where $\wtilde{\y}~=~\y-\1\mbf{\eta}$ when $\mbf{u}(\s,t) \equiv 1$. Next we define an auxiliary variable $\alpha$ that satisfies
\begin{equation*}
    \alpha\geq \wtilde{\y}^\T\covmat^{-1}\wtilde{\y}
\end{equation*}
or equivalently
\begin{equation*}
\begin{bmatrix}
    \alpha & \wtilde{\y}^\T\\
    \wtilde{\y} & \covmat
    \tag{17}
\end{bmatrix} \succeq \0
\end{equation*}
Using the auxiliary variable and the definition of $\covmat$, we can therefore express the objective function as:
\begin{equation*}
    \min_{\alpha,\pvec} ~\alpha + n\theta_0+\sum_{j=1}^{p}\theta_j\|[\basismat]_j\|_2,
    \tag{18}
\end{equation*}
where $\theta_j$ are nonnegative and $\alpha$ satisfies the constraint. The minimizing $\what{\pvec}$ is the learned model parameter. This problem is identified as a convex, semidefinite program, cf. \cite{lobo1998applications}. We may also add the following normalization constraint,
\begin{equation*} 
    \tr\{\samplecovmat - \covmat_\psub \} = 0,
\end{equation*}
to match the normalized covariance matrix. This merely adds a linear constraint to problem (18) with a constrained minimizer denoted $\pvec^\star$. We now prove that $\what{\pvec} \propto \pvec^\star$.

Begin by defining a constant $\kappa > 0$, such that $\text{tr}\{\samplecovmat \covmat^{-1}(\pvec^\star)\} = \kappa^2 \text{tr}\{ \covmat(\pvec^\star)\}$ at the minimum of (18). We show that $\kappa=1$ is the only possible value and so both terms in (18) equal each other at the minimum.

Let $\tilde{\pvec} = \kappa \pvec^\star$, and observe that the cost (18) is then bounded by
\begin{equation*}
\begin{split}
(\kappa^2+1) \text{tr}\{\covmat(\pvec^\star)\} &\leq \text{tr}\{\samplecovmat \covmat^{-1}(\tilde{\pvec} )\} + \text{tr}\{\covmat(\wtilde{\pvec})\} \\
&= \kappa^{-1}\text{tr}\{\samplecovmat \covmat^{-1}(\wtilde{\pvec})\} +\kappa\text{tr}\{ \covmat(\wtilde{\pvec})\}\\
&= 2\kappa \text{tr}\{ \covmat(\wtilde{\pvec})\}.
\end{split}
\end{equation*}
Thus $\kappa$ must satisfy $\kappa^2+1 \leq 2 \kappa$, or $(\kappa-1)^2 \leq 0$. Therefore $\kappa= 1$ is the only solution and both terms must be equal at the minimum. We can thus re-write the minimization of (18) as the following problem
\begin{equation}\label{eq:covariancematch_alt1}
\begin{split}
\min & \quad \alpha \\
\text{subject to} & \quad \text{tr}\{\samplecovmat \covmat^{-1}_\psub\} = \alpha, \; \text{tr}\{ \covmat_\psub \} = \alpha,
\end{split}
\tag{19}
\end{equation}
with minimizer $\what{\pvec}$ and where $\alpha > 0$ is an auxiliary variable.

Next, consider an equivalent problem to (19) obtained by re-defining the variables as $\tilde{\pvec} =  \rho \alpha^{-1}\pvec$. Then $\text{tr}\{\samplecovmat \covmat^{-1}(\pvec)\} = \rho \alpha^{-1} \text{tr}\{\samplecovmat \covmat^{-1}(\tilde{\pvec})\}$ and $\text{tr}\{ \covmat(\pvec) \} = \alpha \rho^{-1} \text{tr}\{ \covmat(\tilde{\pvec}) \}$, so that the equivalent problem becomes
\begin{equation}\label{eq:covariancematch_alt2}
\begin{split}
\min & \quad \beta \\
\text{subject to} & \quad \text{tr}\{\samplecovmat \covmat^{-1}\} = \beta, \; \text{tr}\{ \covmat \} = \rho,
\end{split}
\tag{20}
\end{equation}
where $\beta = \alpha^2 \rho^{-1}$. The minimizer of the equivalent
problem (20) is therefore $\tilde{\pvec}
\propto \pvec^\star$. Problem (20) is
however identical to the constrained problem 
\begin{equation*}
\begin{split}
\min & \quad \text{tr}\{\samplecovmat \covmat^{-1}\}  \\
\text{subject to} & \quad \text{tr}\{ \covmat \} = \rho,
\end{split}
\tag{21}
\end{equation*}
whose minimizer is $\tilde{\pvec} = \what{\pvec}$ when $\rho = \tr\{ \samplecovmat \}$, which follows from expanding the cost in (11) and the normalization constraint. 

Thus we proved that $\what{\pvec} \propto \pvec^\star$ and since the predictor is invariant to uniform scaling of
$\pvec$, that is, $\what{y}_{\what{\psub}}(\s,t) = \what{y}_{\psub^\star}(\s,t)$, we see that the normalization constraint is not relevant for the result.

For further details, see \cite{zachariah2017online}.

\section{\textsc{Equivalent form of the predictor}}

Consider the following augmented problem
\begin{equation*}
\min_{\eta, \: \mbf{v}, \: \pvec} \;  \theta^{-1}_0\| \y -
\1 \mbf{\eta} - \basismat \mbf{v}  \|^2_2 +\| \mbf{v} \|^2_{\pmat^{-1}} + \text{tr}\{
\covmat_\psub \}.
\label{eq:covmatch_aug}
\tag{22}
\end{equation*}
Solving for $\eta$ and $\mbf{v}$ yields the minimizer 
\[
    \mbf{w}^\star = \begin{bmatrix}
    \eta^\star\\
    \mbf{v}^{\star} 
    \end{bmatrix}
    =
    \begin{bmatrix}
        (\1^\T\covmat^{-1}\1)^\dagger\1^\T \covmat^{-1}\y\\
        \pmat\basismat^\T\covmat^{-1}(\y-\1\eta^\star)
        \tag{23}
    \end{bmatrix}.
\]
It can be shown that by inserting the minimizing $\mbf{v}$ back into (22), we obtain a concentrated cost function which is equal to that in (18). Thus we obtain the sought model parameter $\what{\pvec}$ from the augmented problem.

Moreover, we can identify $\mbs{\alpha}^\T(\s, t)\mbf{w}^* = \what{y}_\psub(\s,t)$. Thus we obtain both $\pvec^\star$ and the weights $\mbf{w}^\star$ from the augmented problem. Using these facts, we may alternatively solve for $\pvec$ first. The second and third terms in (22) can be written as
\begin{equation*}
\| \mbf{v} \|^2_{\pmat^{-1}} =  \sum^{p}_{k=1}
\frac{1}{\theta_k}w^2_{1+k}
\end{equation*} 
and
\begin{equation*}
\begin{split} 
\text{tr}\{ \covmat_\psub \} &= \sum^p_{k=1} \|[\basismat]_j\|^2_2 \theta_k + n \theta_0,
\end{split}
\end{equation*}
respectively. Then the minimizing hyperparameters $\pvec$ in \eqref{eq:covmatch_aug} can be expressed in closed-form:
\begin{equation*}
\what{\theta}^{\star}_k = \begin{cases} \|   \y - [\1~~\basismat] \mbf{w}\|_2/\sqrt{n}, \quad k
  = 0.\\
|w_{1+k}|/\|[\basismat]_{1+k}\|_2, \quad k = 1, \dots, p.\end{cases} 
\end{equation*}
Inserting the expression back in to (22)
yields a concentrated cost function
$$\sqrt{\| \y - [\1~~\basismat] \mbf{w} \|^2_2} + \sum^p_{j=1} \frac{1}{\sqrt{n}}\|[\basismat]_j\|_2 |w_{j+1}|$$
which, after dividing by $n^{-1/2}$, equals that in (14). Thus using minimizing weights $\mbf{w}^*$, after concentrating the augmented problem with respect to $\pvec$, yields $\mbs{\alpha}^\T(\s, t)\mbf{w}^* = \what{y}_{\what{\psub}}(\s,t)$

\end{document}